\documentclass[conference]{IEEEtran}
\IEEEoverridecommandlockouts
\usepackage{cite}
\usepackage{amsmath,amssymb,amsfonts}
\usepackage{algorithmic}
\usepackage{graphicx}
\usepackage{textcomp}
\usepackage{xcolor}
\usepackage{multirow}
\usepackage{booktabs} 
\usepackage{bm}
\usepackage{tabularx}
\usepackage{subfigure}
 \usepackage{url}
 \usepackage{colortbl}  %彩色表格需要加载的宏包
\usepackage{xcolor}
\def\BibTeX{{\rm B\kern-.05em{\sc i\kern-.025em b}\kern-.08em
    T\kern-.1667em\lower.7ex\hbox{E}\kern-.125emX}}
\begin{document}

\title{CaDA: Cross-Problem Routing Solver with \underline{C}onstraint-\underline{A}ware \underline{D}ual-\underline{A}ttention\\
% {\footnotesize \textsuperscript{*}Note: Sub-titles are not captured for https://ieeexplore.ieee.org  and
}

\author{\IEEEauthorblockN{1\textsuperscript{st} Han Li}
\IEEEauthorblockA{\textit{School of System Design and}\\ \textit{Intelligent Manufacturing} \\
\textit{Southern University of}\\ \textit{Science and Technology}\\
Shenzhen, China \\
herloconnell@gmail.com}
\and

\IEEEauthorblockN{2\textsuperscript{nd} Fei Liu}
\IEEEauthorblockA{\textit{Department of Computer Science} \\
\textit{City University of Hong Kong}\\
Hong Kong, China \\
fliu36-c@my.cityu.edu.hk}

\and
\IEEEauthorblockN{3\textsuperscript{rd} Zhi Zheng}
\IEEEauthorblockA{\textit{School of System Design and}\\ \textit{Intelligent Manufacturing} \\
\textit{Southern University of}\\ \textit{Science and Technology}\\
Shenzhen, China \\
12010126@mail.sustech.edu.cn}

\and
\IEEEauthorblockN{4\textsuperscript{th} Yu Zhang}
\IEEEauthorblockA{\textit{School of System Design and Intelligent Manufacturing} \\
\textit{Southern University of Science and Technology}\\
Shenzhen, China \\
zhangy5@sustech.edu.cn}
\and
\IEEEauthorblockN{5\textsuperscript{th} Zhenkun
Wang}
\IEEEauthorblockA{\textit{School of System Design and Intelligent Manufacturing} \\
\textit{Southern University of Science and Technology}\\
Shenzhen, China \\
wangzhenkun90@gmail.com}
% \and
% \and
}

\maketitle

\begin{abstract}
Vehicle Routing Problems (VRPs) are significant Combinatorial Optimization (CO) problems holding substantial practical importance. 
% Why NCO
Recently, Neural Combinatorial Optimization (NCO), which involves training deep learning models on extensive data to learn vehicle routing heuristics, has emerged as a promising approach due to its efficiency and the reduced need for manual algorithm design. 
% Demands on Cross-problem learning
However, applying NCO across diverse real-world scenarios with various constraints necessitates cross-problem capabilities. 
% Current NCO methods typically employ a unified model lacking a constraint-specific structure, thereby restricting their cross-problem performance.
Current multi-task methods for VRPs typically employ a constraint-unaware model, limiting their cross-problem performance. Furthermore, they rely solely on global connectivity, which fails to focus on key nodes and leads to inefficient representation learning.
%  
% This paper
This paper introduces a \underline{C}onstraint-\underline{A}ware \underline{D}ual-\underline{A}ttention Model (CaDA), designed to address these limitations. 
CaDA incorporates a constraint prompt that efficiently represents different problem variants. Additionally, it features a dual-attention mechanism with a global branch for capturing broader graph-wide information and a sparse branch that selectively focuses on the most relevant nodes.
% Results
We comprehensively evaluate our model on 16 different VRPs and compare its performance against existing cross-problem VRP solvers. CaDA achieves state-of-the-art results across all the VRPs. Our ablation study further confirms that each component of CaDA contributes positively to its cross-problem learning performance.
\end{abstract}


\section{Introduction}
Vehicle Routing Problems (VRPs) involve optimizing transportation costs for a fleet of vehicles to meet all customers' demands while adhering to various constraints. Numerous studies have focused on VRPs due to their extensive real-world applications in transportation, logistics, and manufacturing~\cite{cattaruzza2017vehicle,rodrigue2020geography}. Traditional methods for solving VRPs include exact solvers and heuristic methods. Exact solvers, however, struggle with the NP-hard nature of the problem, making them prohibitively expensive to implement. On the other hand, heuristic methods are more cost-effective and provide near-optimal solutions but require significant expert input in their design. Recently, learning-based neural solvers have gained considerable attention and have been successfully applied to VRPs~\cite{BENGIO2021405,kool2018attention,bogyrbayeva2024machine}. These solvers train networks to learn a heuristic, reducing the need for extensive manual algorithm design and minimizing computational overhead.

Despite the promising performance of neural solvers on VRPs, the majority of existing works require training a distinct model for each type of routing problem~\cite{NEURIPS2020_f231f210,drori2020learning,duan2020efficiently,gao2020learn,Cappart_2021,9141401,kool2022deep,7969477}. Given that over 60 VRP variants have been studied, each featuring different constraints~\cite{app112110295}, developing distinct models for each routing problem is costly and significantly hinders practical application.

To tackle this challenge, recent efforts have been made to develop cross-problem learning methods that can solve multiple VRPs with a single model~\cite{2024arXiv240216891L,zhou2024mvmoe,berto2024routefinder,Cross_24}. These cross-problem methods typically employ an encoder-decoder framework and are trained using reinforcement learning. For example, MTPOMO~\cite{2024arXiv240216891L} jointly trains a unified model across five VRPs, each with one or two constraints, enabling zero-shot generalization to problems that feature combinations of these constraints. MVMoE~\cite{zhou2024mvmoe} employs a Mixture-of-Experts (MoE) \cite{shazeer2017} structure in the feed-forward layer of a transformer-based model to enhance its cross-problem learning capacity. Furthermore, RouteFinder~\cite{berto2024routefinder} directly trains and tests sixteen VRPs using a proposed unified Reinforcement Learning (RL) environment, which enables the simultaneous handling of different VRPs in the same training batch. Additionally, RouteFinder leverages a modern transformer-based model structure~\cite{dubey2024llama}, along with global embeddings, to enhance performance. 

Despite these advancements, existing cross-problem models remain unaware of constraints \cite{2024arXiv240216891L,zhou2024mvmoe,berto2024routefinder}. As different constraints significantly alter the feasible solution space, this oversight notably limits the models' capabilities in cross-problem applications. 
Furthermore, existing methods employ a transformer encoder which maintains global connectivity throughout the node encoding process, leading to the inclusion of irrelevant nodes and adversely affecting node representation. Conversely, using learnable sparse connections~\cite{chen2023learning,zhou2024adapt} allows the node encoding process to selectively focus on more relevant nodes.

% \fei{% Our work
% This study proposes a novel Constraint-Aware Dual-Attention Model (CaDA) to mitigate this challenge. Firstly, CaDA introduces a constraint prompt to the encoder to enhance the model's awareness of the current problem' constraints. Furthermore, to focus attention scores on particularly promising nodes, we propose a dual-attention structure, which consists of a global branch and a sparse branch with a Top-$k$ operation. While the Top-$k$ operation enables the attention layer to assign exactly zero scores to irrelevant nodes, the global branch with a standard Softmax allows the model to capture the overall graph information. The effectiveness and superiority of the proposed method are comprehensively demonstrated on 16 VRPs. 

% The contributions of this paper can be summarized as follows:
% \begin{itemize} 
% \item We introduce CaDA, an efficient cross-problem learning method for VRPs that enhances model awareness of constraints and prioritizes relevant node pairs.
% \item We propose a constraint-specific prompt and a dual-attention model structure for the CaDA model, which facilitates the extraction of high-quality, focused features for the decoding process.
% \item We conduct a comprehensive evaluation of CaDA across 16 VRP variants. CaDA achieves state-of-the-art performance on all tested variants, surpassing existing cross-problem learning methods. Additionally, our ablation study validates the effectiveness of both the constraint-specific prompt and the dual-attention model. 
% \end{itemize}}

% Our work
This study proposes a novel Constraint-Aware Dual-Attention Model (CaDA) to mitigate these challenges. Firstly, we introduce a constraint prompt to enhance the model's awareness of the activated constraints. Furthermore, we propose a dual-attention mechanism consisting of a global branch and a sparse branch.
% with  Top-$k$ sparse attention. 
% During the decoding process, node pairs with higher attention scores are more likely to be adjacent. Thus, the utilization of sparse branches enables focused learning of representations from these promising nodes.
% During the decoding process, node pairs with higher attention scores are potential neighbors and each customer selects only one next node in the final solution. Since nodes with higher attention scores are more important and promising, the sparse branch enables the model to further refine the learning of relationships between these key nodes.
% \fei{
% The sparse branch allows the model to adaptively identify key node pairs with higher attention scores, thereby enabling the decoder to focus on more promising nodes at each inference step. Meanwhile, the global branch enhances the model's capacity by capturing information from the entire graph, ensuring that the solution is informed globally. The effectiveness and superiority of CaDA have been comprehensively demonstrated across 16 VRPs and real-world benchmarks.}
% During the decoding process, node pairs with higher attention scores are more likely to be to be adjacent in the solution. By incorporating a dual-attention mechanism, the sparse branch targets refining the learning of relationships between these key node pairs, allowing the model to focus on the most promising connections. 
Since, in the encoder-decoder framework, node pairs with higher attention scores are more likely to be adjacent in the solution, the sparse branch with Top-$k$ sparse attention focuses on the more promising connections between these key node pairs.
Meanwhile, the global branch enhances the model's capacity by capturing information from the entire graph, ensuring that the solution is informed globally. 
The effectiveness and superiority of CaDA have been comprehensively demonstrated across 16 VRPs and real-world benchmarks.

The contributions of this paper can be summarized as follows:
\begin{itemize} 
\item We introduce CaDA, an efficient cross-problem learning method for VRPs that enhances model awareness of constraints and representation learning.
\item We propose a constraint prompt, which facilitates high-quality constraint-aware learning, and a dual-attention mechanism, which ensures that the encoding process is both selectively focused and globally informed.
\item We conduct a comprehensive evaluation of CaDA across 16 VRP variants. CaDA achieves state-of-the-art performance, surpassing existing cross-problem learning methods. Additionally, our ablation study validates the effectiveness of both the constraint prompt and the dual-attention mechanism. 
\end{itemize}

\section{Related Work}

\subsection{Neural Combinatorial Optimization} NCO approaches utilize deep reinforcement learning to train a policy that constructs solutions in an autoregressive manner. Nazari et al. \cite{NEURIPS2018_9fb4651c} are the first to apply Pointer Networks \cite{NIPS2015_29921001} to solve the VRP. Subsequently, the pioneering work Attention Model (AM)~\cite{kool2018attention} employs a powerful Transformer-based architecture. This model is optimized using the REINFORCE algorithm~\cite{williams1992simple} with a greedy rollout baseline.
Building on this, Kwon et al. \cite{NEURIPS2020_f231f210} introduce the Policy Optimization with Multiple Optima (POMO) method, which leverages solution symmetries and has demonstrated significantly improved performance. Subsequently, numerous studies have further refined both AM and POMO, enhancing Transformer-based methods~\cite{Xin_Song_Cao_Zhang_2021,NEURIPS2021_564127c0,Dynamic,kim2022sym}. Given the diverse constraints and attributes in real-world transportation needs, some research focuses on various VRP variants, including heterogeneous Capacitated VRP (HCVRP) \cite{9547060}, VRP with Time Windows (VRPTW)~\cite{gao2020learn,Cappart_2021,9141401, kool2022deep}, and Open Route VRP (OVRP)~\cite{7969477}. More information can be found in recent reviews \cite{bogyrbayeva2022learning,li2022overview}.

\subsection{Cross-Problem Learning for VRPs} 
Neural methods for solving VRPs typically train and evaluate deep models on the same instance distributions. Some studies have explored generalization across multiple distributions \cite{zhang2022learning, xin2022generative,geisler2022generalization, wang2022a, Jiang_Wu_Cao_Zhang_2022, NEURIPS2022_ca70528f}. Additionally, Zhou et al. \cite{zhou2023towards} consider both problem size and distribution variations. Recent developments have begun to address cross-problem generalization \cite{2023arXiv230506361W, Cross_24,2024arXiv240216891L, zhou2024mvmoe}.  
Wang and Yu \cite{2023arXiv230506361W} use multi-armed bandits to achieve task scheduling. Lin et al. \cite{Cross_24} demonstrate how a model pre-trained on the Travelling Salesman Problem (TSP) could be effectively adapted to targeted VRPs through efficient fine-tuning, e.g., inside tuning, side tuning, and Low-Rank Adaptation (LoRA). 
% UNCO~\cite{jiang2024unco} uses natural language to formulate CO instances and employs a frozen Large Language Model (LLM) to encode them. The resulting embeddings are then processed through a multi-layer attention block to extract node embeddings. However, the LLM is input-sensitive, requiring expensive manual adjustments to instance formulations. Additionally, when the LLM and the multi-layer attention block undergo separate training, the resulting solutions are often sub-optimal. 
However, these approaches still focus on relatively few tasks (less than five).

To tackle multiple VRP variants in a unified model, MTPOMO \cite{2024arXiv240216891L} conceptualizes VRP variants as combinations of underlying constraints, enabling the model to achieve zero-shot generalizability to more tasks. MVMoE proposes a new model architecture using the MoE \cite{shazeer2017} approach to improve performance. Furthermore, RouteFinder~\cite{berto2024routefinder} proposes to use a modern transformer encoder structure incorporating SwiGLU~\cite{dauphin2017language}, Root Mean Square Normalization (RMSNorm)~\cite{zhang2019root}, and Pre-Norm~\cite{baevski2018adaptive, child2019generating}, which considerably improves the model's capability. However, these approaches remain unaware of constraints and maintain only global connectivity throughout the encoding process, which limits their cross-problem capabilities.

\subsection{Multi-Branch Architecture}
Multi-branch architectures have been widely used and have achieved success in computer vision. Some research employs multiple branches to capture both low and high-resolution image information, ultimately producing a comprehensive and powerful semantic representation that can be used for downstream tasks such as image segmentation~\cite{ronneberger2015u,guo2022segnext,gu2022multi} or human pose estimation~\cite{sun2019high,wang2020deep}. Other studies assign different branches to focus on distinct aspects by utilizing attention mechanisms~\cite{fu2019dual,zhang2022mixste,dong2022cswin,wang2023crossformer++}. For instance, DANet~\cite{fu2019dual} proposes a dual attention network for scene segmentation, with one branch responsible for capturing pixel-to-pixel dependencies and another for capturing channel dependencies across different feature maps, thereby capturing global context~\cite{zhang2018context}. Similarly, Crossformer++~\cite{wang2023crossformer++} groupes image patches in both local and global ways, incorporating short-distance and long-distance attention to achieve better representation, retaining both small-scale and large-scale features in the embeddings. 
In this study, we propose a novel dual-attention mechanism featuring a global branch for capturing information from the entire graph and a sparse branch to focus more on important connections.

\subsection{Sparse Attention}
% Recently, many studies have introduced fixed or learnable sparsity patterns, such as local or strided attention \cite{guo2019star, beltagy2020longformer, ainslie2020etc}, enabling each token to attend only to a subset of tokens~\cite{child2019generating, tay2020sparse,}.
Recent studies have proposed using sparse attention to reduce computational complexity and minimize the harmful influence of unnecessary and irrelevant items, thereby improving performance~\cite{zhao2019explicit, wang2022kvt, wang2022nformer, chen2023learning, zhao2023comprehensive}. 
To achieve this,
% attention weights of exactly zero, 
many researchers utilize pre-defined sparse attention patterns based on prior knowledge, such as local or strided attention, or combinations of multiple patterns~\cite{guo2019star, li2019enhancing, beltagy2020longformer, ainslie2020etc}. For instance, LogSparse~\cite{li2019enhancing} ensured that each token only attends to itself and its preceding tokens, using an exponential step size. However, these methods can be overly harsh and require well-informed prior knowledge. 
Another category of methods achieves sparse attention by adding an additive operation that eliminates small attention scores to exactly zero, such as the Top-$k$ operation \cite{zhao2019explicit,wang2022kvt,wang2022nformer,chen2023learning} and the $\text{ReLU}^2$ operation~\cite{zhou2024adapt}, or employs a sparsity-inducing alternative to Softmax, such as sparsemax~\cite{martins2016sparsemax} and $\alpha$-entmax~\cite{peters2019entmax,blondel2019learning,correia2019adaptively}. In this study, we use the simple yet efficient Top-$k$ selection operation to achieve sparse attention and enhance the representation learning from the most relevant nodes.

\section{Preliminaries}
\subsection{Problem Definition}

\begin{figure}[t]
\centering
\includegraphics[width=0.9\linewidth]{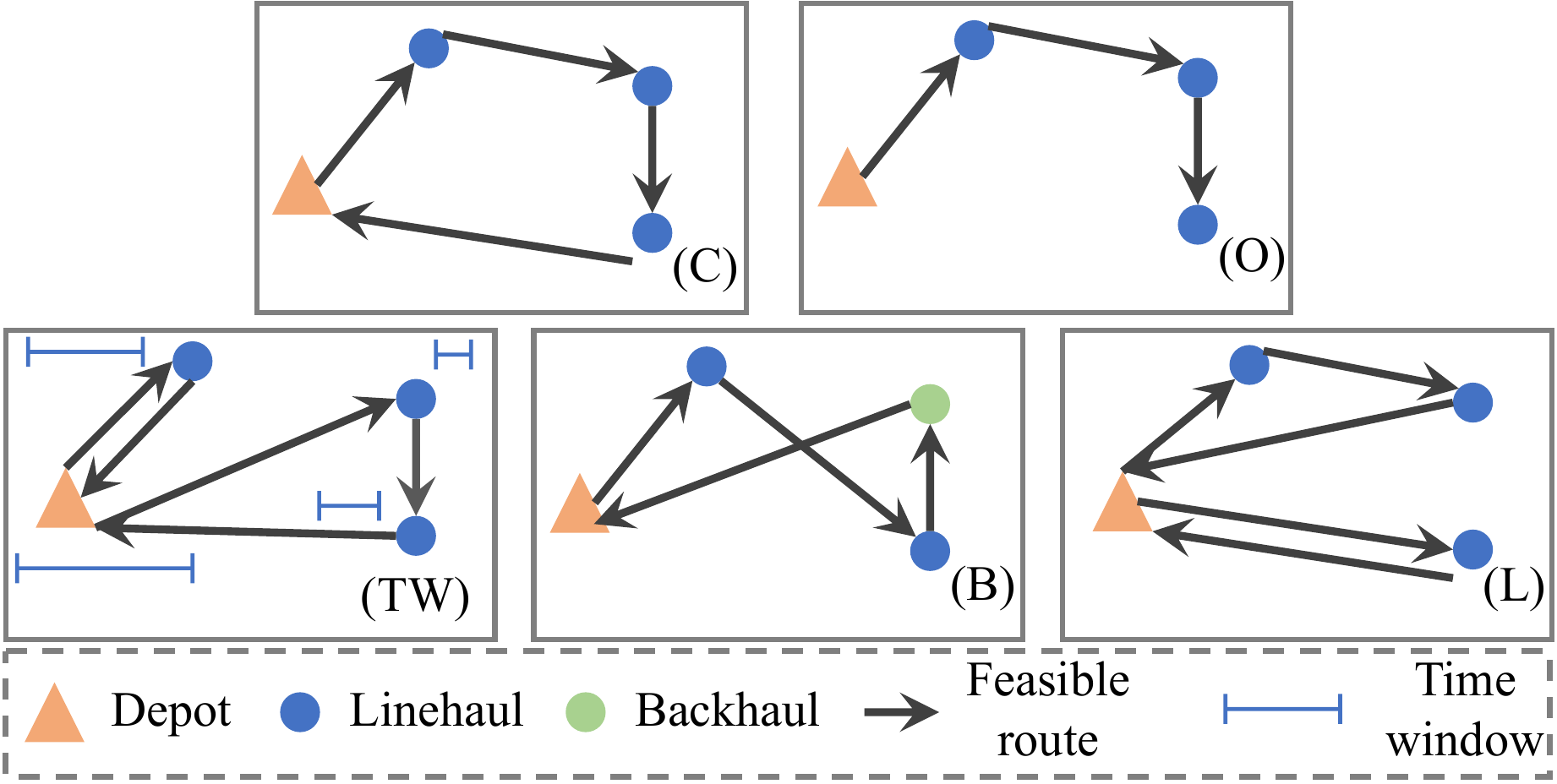}
\caption{The feasible solutions for different VRP variants.}
\label{fig:5c}
\end{figure} 

\begin{table}[t]
  % \vskip -0.2in
  \caption{16 VRP variants with five constraints.}
  \label{tab:16p}
  % \vskip 0.1in
  \begin{center}
  % \begin{small}
  % \renewcommand\arraystretch{1.0} % 调整行高
  \resizebox{1.\columnwidth}{!}{  % 调整宽度
  \begin{tabular}{l|ccccc}
    \toprule
    % \midrule
    % & Capacity (C) & Open Route (O) & Backhaul (B) & Duration Limit (L) & Time Window (TW) \\
    % & C & O & B & L & TW \\
    & Capacity & Open Route & Backhaul & Duration Limit & Time Window \\
    \midrule
    CVRP & \checkmark & & & & \\
    % \cmidrule{2-6}
    OVRP & \checkmark & \checkmark & & & \\
    VRPB & \checkmark & & \checkmark & & \\
    VRPL & \checkmark & & & \checkmark & \\
    VRPTW & \checkmark & & & & \checkmark \\
    OVRPTW & \checkmark & \checkmark & & & \checkmark \\
    OVRPB & \checkmark & \checkmark & \checkmark & & \\
    OVRPL & \checkmark & \checkmark & & \checkmark & \\
    VRPBL & \checkmark & & \checkmark & \checkmark & \\
    VRPBTW & \checkmark & & \checkmark & & \checkmark \\
    VRPLTW & \checkmark & & & \checkmark & \checkmark \\
    OVRPBL & \checkmark & \checkmark & \checkmark & \checkmark & \\
    OVRPBTW & \checkmark & \checkmark & \checkmark & & \checkmark \\
    OVRPLTW & \checkmark & \checkmark & & \checkmark & \checkmark \\
    VRPBLTW & \checkmark & & \checkmark & \checkmark & \checkmark \\
    OVRPBLTW & \checkmark & \checkmark & \checkmark & \checkmark & \checkmark \\
    % \midrule
    \bottomrule
  \end{tabular}
  }
%   \end{sc}
  % \end{small}
  \end{center}
  % \vskip -0.1in
\end{table}

In this study, we focus on 16 VRP variants that encompass five different constraints, including Capacity (C), Open Route (O), Backhaul (B), Duration Limit (L), and Time Window (TW). These are summarized in Table~\ref{tab:16p}, with illustrations related to these constraints presented in Figure~\ref{fig:5c}. In this section, we begin by outlining a general definition of the VRP instance, then continue with the basic CVRP, and proceed to describe its four additional constraints.

A VRP instance $\mathcal{G}$ is a fully-connected graph defined by a set of nodes $\mathcal{V} = \{ v_0, v_1, \ldots, v_{N} \}$ with the total number of nodes given by $|\mathcal{V}| = N+1$, and edges $\mathcal{E} = \mathcal{V} \times \mathcal{V}$. Furthermore, $v_0$ represents the depot, while $\{v_1, \ldots, v_{N}\}$ represent the $N$ customer nodes. Each node $v_i \in \mathcal{V}$ consists of the pair $\{ \vec{X}_i, A_i\}$, where $\vec{X}_i \sim U(0,1)^2$ represents the node coordinates, and $A_i$ denotes other attributes of the nodes. Additionally, the travel cost between different nodes is defined by their Euclidean distance, which is denoted by the cost matrix $\mathbf{D} = \{ d_{i,j}, i=0, \ldots, N, j=0\ldots, N\}$.

In CVRP, the depot node $v_0$ has $A_0 = \emptyset$, and each customer node $v_i$ is associated with $A_i = \{ \delta_{i} \}$, where $\delta_{i}$ is the customer's demand $v_i$ that the fleet of vehicles must service. This fleet comprises homogeneous vehicles, each with a specific capacity $C$. Each vehicle leaves the depot $v_0$, visits a subset of customers, and returns to the depot upon completion of deliveries. 
% The solution to the CVRP, denoted by $ \boldsymbol{\tau} $, consists of the routes taken by all vehicles. The objective is to minimize the total travel distance for all vehicles, constrained by ensuring that each customer is visited exactly once and that the total demand carried on each route does not exceed the vehicle's capacity. The objective function for the CVRP can be mathematically defined as:
The solution to CVRP, denoted by $\boldsymbol{\tau}$, consists of the routes taken by all vehicles, that is, $\boldsymbol{\tau} = \{\boldsymbol{\tau}^{1}, \boldsymbol{\tau}^{2}, \ldots, \boldsymbol{\tau}^{K}\}$, where $K$ is the total number of sub-routes. Each sub-route $\boldsymbol{\tau}^{k} = (\tau^{k}_1, \tau^{k}_2, \ldots, \tau^{k}_{n_k})$, $k \in {1, 2, \ldots, K}$, where $\tau^{k}_i$ is the index of the visited node at step $i$, and $\tau^{k}_1 = \tau^{k}_{n_k} = 0$. $n_k = |\boldsymbol{\tau}^k|$ represents the number of nodes in it, and $\sum_{k=1}^{K}n_{k} = T$ represents the total length of the solution.

% \begin{equation}
%     \begin{aligned}
%         \mbox{minimize}\quad & f(\boldsymbol{\tau})=\sum_{k=1}^{K}F(\boldsymbol{\tau}^{k}), \\
%         & F(\boldsymbol{\tau}^{k})=\displaystyle \sum_{t=0}^{n_k-1}d_{\tau^{k}_{t},\tau^{k}_{t+1}}+d_{\tau^{k}_{n_{k}}, \tau^{k}_{0}},\\
%         \mbox{subject to}\quad & 0 \leq \delta_{i} \leq C, \quad i=1,\ldots,N,\\
%        &\displaystyle \sum_{i \in \boldsymbol{\tau}^{k}} \delta_{i} \leq C, \quad k=1,\ldots,K,
%     \end{aligned}
% \end{equation}
% where $\boldsymbol{\tau}$ consists of $K$ sub-routes $\boldsymbol{\tau} = \{\boldsymbol{\tau}^{1}, \boldsymbol{\tau}^{2}, \ldots, \boldsymbol{\tau}^{K}\}$. Each sub-route $\boldsymbol{\tau}^{k} = (\tau^{k}_1, \tau^{k}_2, \ldots, \tau^{k}_{n_k})$, $k \in \{1, 2, \ldots, K\}$, must start from the depot $v_{0}$ and return to $v_{0}$ in the end. Here, $n_k = |\boldsymbol{\tau}^k|$ represents the number of customer nodes in it, and $\sum_{k=1}^{K}n_{k} = T$ represents the total length of the solution.

The basic CVRP could be easily extended to accommodate various VRPs by incorporating additional constraints. In this paper, we explore four additional constraints as discussed in recent studies~\cite{2024arXiv240216891L,zhou2024mvmoe,berto2024routefinder}. 

\paragraph{ Open Route (O)} In the OVRP, vehicles do not return to the depot $v_0$ after completing their sub-route. 

\paragraph{ Time Window (TW)} The Time Window constraint requires that each node must be visited within a specific time window, such that each node $A_i = \{\delta_{i}, e_i, l_i, s_i \}$, where $e_i$ is the earliest start time, $l_i$ is the latest permissible time, and $s_i$ represents the time taken to service this customer. The depot $v_0$ has $s_0=0$, $e_0=0$, and $l_0=\mathcal{T}$, indicating that each sub-tour must be completed within a time limit of $\mathcal{T}$. Time window constraints are stringent; if a vehicle arrives earlier than $e_i$, it must wait until the start of the window.

\paragraph{ Backhaul (B)} Customers with $\delta_i > 0$ are termed linehaul customers, as they require vehicles to load goods at the depot and deliver to their locations. Conversely, customers with $\delta_i < 0$ are defined as backhaul customers, where vehicles must collect $|\delta_i|$ from their locations and transport it back to the depot. While all customers in the standard CVRP are linehaul, the Vehicle Routing Problem With Backhauls(VRPB) includes both linehaul and backhaul customers. Furthermore, all linehaul tasks must be completed before any backhaul tasks can commence on the route to avoid rearranging the loads on the vehicle.

\paragraph{Duration Limit (L)} In this constraint, the depot $v_0$ has $A_0 = \{ \rho \}$, where $\rho$ represents the length limit that each sub-tour must adhere to, ensuring a return to $v_0$ within this threshold.

\subsection{Learning to Construct Solutions for VRPs} 
The process of constructing solutions autoregressively (i.e., during decoding) can be modeled as a Markov Decision Process (MDP), and the policy can be trained using RL methods. As the model sequentially expands each sub-route, for simplicity, at any decoding step $t$, $\boldsymbol{\tau}_t$ represents the sequence of nodes visited up to that point:
\begin{equation}
    \boldsymbol{\tau}_t = \bigcup\limits_{k=1}^{K} (\tau^{k}_1, \tau^{k}_2, \ldots, \tau^{k}_{n_k}) = (\tau_1, \tau_2, \ldots, \tau_t),
\end{equation}
where $\bigcup$ denotes the concatenation of sequences from different sub-routes
% , and $\tau_i, 1 \leq i \leq t$, and $\tau_j, 1 \leq j \leq t$, may  belong to different sub-routes. 
The MDP for the decoding step $t$ can be defined as follows:

\paragraph{State} $s_t \in \mathcal{S}$ is the ordered tuple $(\boldsymbol{\tau}_{t-1}, \mathcal{V})$ given by the current partial solution $\boldsymbol{\tau}_{t-1} = (\tau_1, \tau_2,\ldots, \tau_{t-1})$ and the instance $\mathcal{V}$. Initially, $\boldsymbol{\tau}_0 = \emptyset$, and at the end, $s_T$ contains a feasible solution $\boldsymbol{\tau}_T$.  

\paragraph{Action} $a_t \in \mathcal{A}$ is the selected index in the current step, which will be added at the end of the partial solution. If $a_t = 0$, i.e., the vehicle returns to the depot node, it signifies the end of the current sub-tour and the start of a new one. %: $s_{t+1} = ((0), \mathcal{V})$.

\paragraph{Policy} A neural model $\pi_{\theta}$ with learnable parameters $\theta$ is used as a policy to generate solutions sequentially, where the probability of generating the final feasible solution is:
\begin{align}
\pi_\theta(\boldsymbol{\tau}|\mathcal{V}) &= \prod_{t=1}^{T} \pi_\theta(a_t|s_t) = \prod_{t=1}^{T} \pi_{\theta}(\tau_{t} \mid \boldsymbol{\tau}_{t-1}, \mathcal{V}).
\end{align}

\paragraph{Reward} $r \in \mathcal{R}$ can only be obtained when a whole feasible solution $\boldsymbol{\tau}_T$ is generated and is defined as the negative solution length:

\begin{equation}
    \begin{aligned}
r(\boldsymbol{\tau}_{T}) &= - \displaystyle \sum_{t=1}^{T-1}d_{\tau_{t}\tau_{t+1}}.
    \end{aligned}
\end{equation}

Subsequently, $\pi_{\theta}$ can be optimized using RL methods to maximize the expected reward $J$. This study employs the REINFORCE algorithm \cite{williams1992simple} with a shared baseline proposed by Kwon et al. \cite{NEURIPS2020_f231f210}, to update the policy. Specifically, for a VRP instance $\mathcal{V}$, $N$ trajectories are generated, starting with the first action $\{a_1^1, a_1^2, \dots, a_1^N\}$, which is always $0$. 
% Each of the $N$ trajectories then assigns a unique one of the $N$ customer nodes, $\{v_0, v_1, \dots, v_N\}$, as the second point. 
Each of the $N$ trajectories then assigns a unique one of the $N$ customer nodes as the second point, i.e., $\{a_2^1, a_2^2, \dots, a_2^N\} = \{1, 2, \dots, N\}$.
The policy subsequently samples actions for each trajectory until all have derived feasible solutions $\{\boldsymbol{\tau}^1, \boldsymbol{\tau}^2, \dots, \boldsymbol{\tau}^N\}$. Finally, the gradient of the policy is approximated by:
\begin{equation}
\begin{aligned}
\nabla_\theta J(\theta \mid \mathcal{V}) &\approx \frac{1}{N}\sum_{i=1}^N (r(\boldsymbol{\tau}^i) - b^i(\mathcal{V})) \nabla_{\theta} \log \pi_{\theta}(\boldsymbol{\tau}^i \mid \mathcal{V}), \\ 
b^i(\mathcal{V}) &= \frac{1}{N} \sum_{j=1}^{N} r(\boldsymbol{\tau}^j) \quad \textrm{for all} \enspace i.
\end{aligned}
\end{equation}
Where $b(\mathcal{V})$ is the shared baseline function used to stabilize learning. 

For the structure of policy $\pi_{\theta}$, existing approaches primarily use transformer-based models.

\subsection{Transformer Layer} 
The Transformer \cite{vaswani2017attention} comprises a Multi-Head Attention Layer (MHA) and a Feed-Forward Layer (FFD). In some modern large language models~\cite{chowdhery2023palm, touvron2023llama, naveed2023comprehensive}, the FFD is replaced by Gated Linear Units (GLUs).
\paragraph{Attention Layer} 
The classical attention function can be expressed as:
% \begin{align}
% \text { Attention }(X,Y)=\operatorname{Softmax}\left(\frac{X W_Q (Y W_K)^{\top}}{\sqrt{d_{k}}}\right) Y W_V,
% \end{align}
\begin{equation}
\begin{aligned}
\label{eq:attn}
\text { Attention }(X,Y) & = \mathbf{A} \left(Y W_V \right), \\
\text{where } \mathbf{A} &= \operatorname{Softmax}\left(\frac{X W_Q (Y W_K)^{\top}}{\sqrt{d_{k}}}\right),\\
\end{aligned}
\end{equation}
where $X \in \mathbb{R}^{n \times d}$ and $ Y \in R^{m\times d}$ represent the input embeddings. The parameters $W_Q, W_K \in \mathbb{R}^{d \times d_k}$, and $W_V \in \mathbb{R}^{d \times d_v}$ are trainable matrices for the query, key, and value projections, respectively. 
After calculating the attention matrix using the query and key matrices, the Softmax function is applied independently across each row to normalize the attention scores. These scores are then rescaled by $\sqrt{d_k}$, resulting in the scaled attention score matrix $\mathbf{A}$. The eventual output, denoted as $Z$, is a matrix in $\mathbb{R}^{n \times d_v}$. 

Additionally, for efficiency, the MHA projects $X$ into $M_h$ separate sets of queries, keys, and values, upon which the attention function is applied:
\begin{equation}
\begin{aligned}
    &\text{MHA}(X,Y)=\text{Concat}(Z_1,\ldots,Z_{M_h})W_p,\\
    &\quad \text{where}\quad Z_i=\text{Attention}_i(X,Y), \forall i\in\{1,\ldots,M_h\},
\end{aligned}
\label{mha}
\end{equation}
where $d_k = d_v = \frac{d}{M_h}$ in each $\text{Attention}_i$. The parameter $W_p \in \mathbb{R}^{d \times d}$ denotes a trainable matrix combining different attention heads' outputs. For self-attention, we have $Y = X$.
% meaning the attention function can be written as $\text{MHA}(X, X)$. For simplicity and without losing accuracy, we can refer to it as $\text{MHA}(X)$. 

\begin{figure*}[t]
\centering
\includegraphics[width=1.\linewidth]{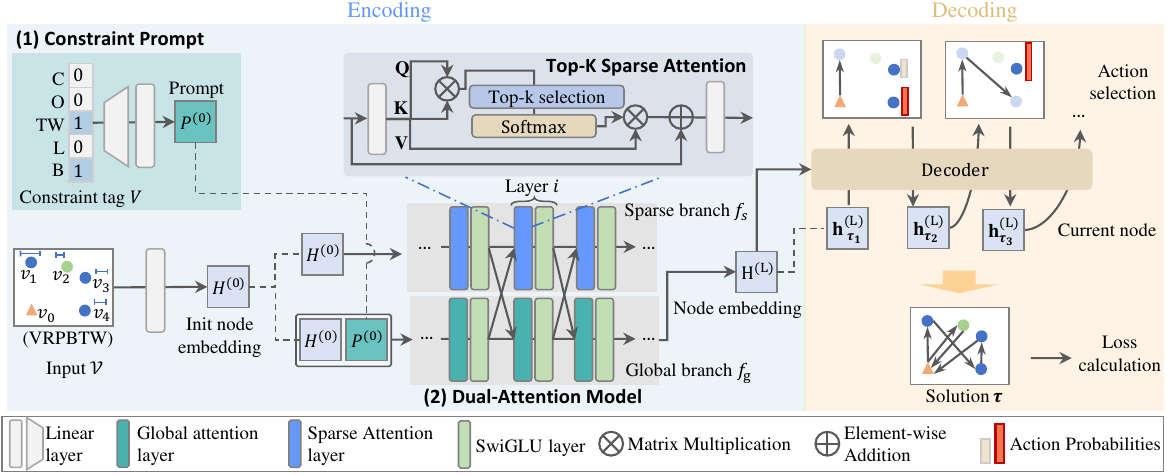}
\caption{The pipeline of the proposed CaDA for VRPs. CaDA adopts the typical encoder-decoder framework and incorporates two new components in the encoder: a dual-attention mechanism and a constraint prompt. The dual-attention mechanism comprises a global branch with the standard Softmax function and a sparse branch with Top-$k$ operation.
}
\label{fig:framework_}
\end{figure*}

\paragraph{Gated Linear Unit} The transformer blocks also include an FFD that processes the input $X$ through two learned linear projections, with a ReLU activation function applied between them. In many recent modern transformer-based large language models~\cite{chowdhery2023palm, touvron2023llama, naveed2023comprehensive}, this configuration has been replaced by GLUs~\cite{dauphin2017language}. GLUs consist of a component-wise product of two linear projections, where one projection is first passed through a nonlinear function. We employ SwishGLU~\cite{dauphin2017language}, which utilizes the Sigmoid Linear Unit (SiLU)~\cite{elfwing2018sigmoid} as the nonlinear function, as recommended in the RouteFinder\cite{berto2024routefinder}. The SwiGLU is defined as:
\begin{align} 
\text{SwiGLU}(X) &= X \odot \sigma(XW_1 + b_1) \otimes \text{SiLU}(XW_2 + b_2).
\end{align}
where $\odot$ denotes element-wise multiplication, $\otimes$ is matrix multiplication, $\sigma$ is the sigmoid function, and $W_1, W_2, b_1, b_2$ are learnable parameters.

\section{Methodology}

\subsection{Overall Pipeline}
As shown in Figure~\ref{fig:framework_}, CaDA follows the general cross-problem learning framework for VRPs which consists of two stages: encoding instance $\mathcal{V}$ to node embeddings $H^{(L)}$, and decoding to construct solutions based on $H^{(L)}$ sequentially. CaDA employs a prompt to introduce constraint information during the encoding process and further utilizes a dual-attention mechanism to enhance representation learning.

\subsection{Constraint Prompt} 
To generate prompts that carry the problem's constraint information, we represent the problem as a multi-hot vector $ V \in \mathbb{R}^5 $, corresponding to five distinct constraints. This multi-hot vector is subsequently processed through a straightforward Multi-Layer Perceptron (MLP) to generate the prompts:
\begin{align}
P^{(0)} = \text{LayerNorm}(V W_a + b_a) W_b + b_b,
\end{align}
where $W_a \in \mathbb{R}^{5 \times d_h}$, $b_a \in \mathbb{R}^{d_h}$, $W_b \in \mathbb{R}^{d_h \times d_h}$, and $b_b \in \mathbb{R}^{d_h}$ are learnable parameters. $d_h$ is the node's embedding dimension. Then this prompt can be concatenated with the node embeddings.

\subsection{Dual-Attention Mechanism}
The input instance $\mathcal{V} $ with $ |\mathcal{V}| = N+1$, is first transformed into high-dimensional initial node embeddings a linear projection. The initial node embedding is denoted as $H^{(0)} \in \mathbb{R}^{(N+1) \times d_h}$. 
% Additionally, multi-hot constraint vectors are used to generate initial prompts $P^{(0)}$.

Subsequently, $H^{(0)}$ is concatenated with $P^{(0)}$ and processed through a global branch $f_g$, which consist of $L$ layers. Each consists of a standard MHA layer~\cite{vaswani2017attention} and a SwiGLU~\cite{shazeer2020glu}. The standard attention function with Softmax never allocates exactly zero weight to any node, thereby allowing each node access to the entire graph. Concurrently, to capture information from closely related nodes, a sparse branch denoted as $f_s$ with Top-$k$ sparse attention layers is introduced. Both branches adaptively fuse information at the end of each layer.

Finally, the output from the global branch, $H^{(L)}_g$, is used for autoregressive decoding, with the likelihood of node selection being primarily determined by the similarity of the nodes' embeddings.

% \cite{vaswani2017attention} and a feed-forward (FF) layer, enhanced by instance normalization (IN) \cite{ulyanov2016instance} and residual connections \cite{he2016deep}. Additionally, branch \( f_b \) concatenates the output from the instance branch and applies an additional linear projection to fuse this information.

\paragraph{Global Layer} 
Each layer involves a MHA \cite{vaswani2017attention} and a SwiGLU~\cite{shazeer2020glu}, along with RMSNorm \cite{zhang2019root} and residual connections \cite{he2016deep}. The $i$-th layer is formulated as follows:
\begin{align}
\hat{H}_{g}^{(i)} &= \text{RMSNorm}^{(i)}\left(H_{g}^{(i-1)} + \text{MHA}^{(i)} \right. \nonumber \\
&\quad \left. \left(H_{g}^{(i-1)}, \text{Concat}\big[H_{g}^{(i-1)}, P^{(i-1)} \big]\right)\right), \\
\tilde{H}_{g}^{(i)} &= \text{RMSNorm}^{(i)}\left(\hat{H}_{g}^{(i)} + \text{SwiGLU}^{(i)}(\hat{H}_{g}^{(i)})\right), \\
\hat{P}^{(i)} &= \text{RMSNorm}^{(i)}\left(P^{(i-1)} + \text{MHA}^{(i)} \right. \nonumber \\
&\quad \left. \left(P^{(i-1)}, \text{Concat}\big[H_{g}^{(i-1)}, P^{(i-1)} \big]\right)\right) , \\
{P}^{(i)} &= \text{RMSNorm}^{(i)}\left(\hat{P}^{(i)} + \text{SwiGLU}^{(i)}(\hat{P}^{(i)})\right),
\end{align}
where ${H}_{g}^{(i-1)} \in \mathbb{R}^{(N+1) \times d_h}$ represents the node embeddings output from the $(i-1)$-th global layer.

\paragraph{Sparse Layer}
In the sparse branch $f_s$, each layer also consists of an attention layer and a SwiGLU activation function. However, to focus more precisely on related nodes, we replace the attention function $\text{Attention}(\cdot,\cdot)$ in $\text{MHA}(\cdot,\cdot)$ with $\text{SparseAtt}(\cdot, \cdot)$, which masks attention scores smaller than the Top-$k$ scores by setting them to zero. This can be formulated as follows:
\begin{equation}
   \text{SparseAtt}(X,Y)=\operatorname{Softmax}\left({M}( \mathbf{A} ) \right) YW_V,
\end{equation}
where $\mathbf{A}$ is the attention scores calculated as shown in Equation~\ref{eq:attn}. ${M}(\cdot)$ is the Top-$k$ selection operation:
\begin{equation}
    \left[{M}( \mathbf{A} )\right]_{ij} = \begin{cases} 
     \mathbf{A} _{ij} & \text{if }  \mathbf{A} _{ij} \in \text{Top-}k( \mathbf{A} _{i*}), \\
    0 & \text{otherwise} .
    \end{cases}
\end{equation}
where $\mathbf{A}_{i*}$ represents the attention scores of the $i$-th node with all other nodes, i.e., $\mathbf{A}_{i*} = \{ \mathbf{A}_{ij} \ | \ j \in \{0, 1, \dots, N\} \}$, and the Top-$k$ operation selects the top $k$ highest attention scores from this set.

\paragraph{Fusion Layer}
In our model, a simple linear projection is applied at the end of each layer to transform embeddings between two branches. For the $i$-th layer, the outputs from the global and sparse branches are denoted as $\tilde{H}^{(i)}_{g}$ and $\tilde{H}^{(i)}_{s}$, respectively. The final outputs are given by:
\begin{align} 
H^{(i)}_{g} &= \tilde{H}^{(i)}_{g} + (\tilde{H}^{(i)}_{s} W_s + b_s), \\ H^{(i)}_{s} &= \tilde{H}^{(i)}_{s} + (\tilde{H}^{(i)}_{g} W_g + b_g), \end{align}
where $W_s$, $b_s$, $W_g$, and $b_g$ are learnable parameters.
\subsection{Decoder}
After encoding, the output of the global branch, $H^{(L)} = [\boldsymbol{h}_{0}^{(L)}, \boldsymbol{h}_{1}^{(L)}, \ldots, \boldsymbol{h}_{N}^{(L)}]$, is utilized to construct the solution. During the autoregressive decoding process, at step $t$, the context embedding is defined as:
\begin{equation}
        {H}_{c}  =\text{Concat}\big[\boldsymbol{h}^{(L)}_{{\tau}_{t}},c_{t}^{\text{l}}, c_{t}^{\text{b}}, z_{t}, l_{t}, o_{t} \big] W_{t},
\end{equation}
where $\boldsymbol{\tau}_{t}$ is the partial solution already generated, and ${\tau}_{t}$ is the last node of the partial solution. The terms $c_{t}^{\text{l}}, c_{t}^{\text{b}}$ represent the remaining capacity of the vehicle for linehaul and backhaul customers, respectively. The terms $z_t$, $l_t$, and $o_t$ represent the current time, the remaining length of the current partial route (if the problem includes a length limitation), and the presence indicator of the open route, respectively. The matrix $W_c \in \mathbb{R}^{(d_h + 5) \times d_h}$ is a learnable parameter.

Then the context embeddings are processed through a MHA to generate the final query: 
\begin{equation}
    \begin{aligned}
    {q}_{c}=\text{MHA}(H_{c}^{(L)},\text{Concat}\big[\boldsymbol{h}_i^{(L)} : i \in I_t\big]),
    \end{aligned}
\end{equation}
where $I_t$ represents the set of feasible actions at the current step.
The compatibility \(u_i\) is computed as:
\begin{align}
u_{i} = \begin{cases} 
\xi \cdot \tanh\left(\frac{{q}_{c} (\boldsymbol{h}_i^{(L)})^{\top}}{\sqrt{d_k}}\right) & \text{if } i \in I_t , \\
-\infty  & \text{otherwise},\\
\end{cases}
\end{align}
where $\xi$ is a predefined clipping hyperparameter. 
Finally, the action probabilities $\pi_{{\theta}}({\tau}_{g} = i \mid \mathcal{V}, \boldsymbol{\tau}_{1:g-1})$ are obtained by applying the Softmax function to $u = \{u_i\}_{i \in I}$.

Additionally, to determine the set of feasible actions at the current step $I_t$, we utilize the following feasibility testing process.

\begin{enumerate}
    \item Each customer node can only be visited once. If the depot is the last action in a partial solution, the next action cannot be the depot (to avoid a self-loop).
    \begin{align}
         i \in \boldsymbol{\tau}_{1:t-1}, i \in \{1, 2, \dots, N\} \quad \Rightarrow \quad i \notin I_t, \\
        {\tau}_{t-1}=0 \quad \Rightarrow \quad 0 \notin I_t.         
    \end{align}
    \item For a problem without the Open Route constraint ($V_1 = 0$), each sub-route needs to return to the depot $v_0$ within the given limit. There are two types of constraints that enforce limits on when each sub-route must reach the depot: the Time Window constraint ($V_2 = 1$) with a time limit $\mathcal{T}$, and the Distance Limit constraint ($V_3 = 1$) with a distance limit $\rho$.
    \begin{align}
    &(V_1 = 0) \land (V_2 = 1) \land \left((z_t + d_{{\tau}_{t-1}i}+s_i+d_{i0}) > \mathcal{T}\right) \nonumber  \\
    &\vee  \left((V_1 = 0) \land (V_3 = 1) \land \left(l_t < d_{{\tau}_{t-1}i}+d_{i0}\right)\right) \nonumber \\
    &\Rightarrow i \notin I_t.
    \end{align}
    \item For problems with a Time Window constraint ($V_2 = 1$), each customer node $v_i$ has a time window $[e_i, l_i]$ and a service time $s_i$. The vehicle must visit $v_i$ and complete its service within the specified time window.
    \begin{equation}
    (V_2 = 1) \land (z_t +d_{\tau_{t-1}i}+s_i > l_i)  \Rightarrow i \notin I_t. 
    \end{equation}
    \item For the problem with the Backhaul constraint ($V_4 = 1$), the backhaul will be masked if there are still linehaul services that have not been completed.
    \begin{align}
    &\left( \delta_i < 0 \right) \land \left( \exists j\in \{1,2,\dots,N \} (\delta_j > 0) \land (j \notin \tau_{1:t-1}) \right) \nonumber \\
    &\Rightarrow i \notin I_t.
    \end{align}
    \item For customers, service is available when their demand does not exceed the current available capacity.
    \begin{align}
    &\left((\delta_i > 0) \land  (\delta_i > c_t^{\text{l}})\right) \lor 
    \left( (\delta_i < 0) \land  (-\delta_i > c_t^{\text{b}}) \right) \nonumber \\
    &\Rightarrow i \notin I_t . 
    \end{align}

\end{enumerate}

%%%% IF 没有被上面的条件

\section{Experiments}
To evaluate the effectiveness of the proposed CaDA for VRPs, we conduct experiments on 16 different VRP variants with five constraints. Furthermore, we perform ablation studies to validate the efficiency of the proposed components.

\subsection{Problem Setup} 

In this section, we provide a detailed description of the VRP problems setup used in this study.

\paragraph{Locations} The nodes' locations are represented by a two-dimensional vector $\vec{X}_i, i \in \{0,..
.,N\}$, and are derived from a uniform distribution $U(0,1)$.

\paragraph{Capacity} In this study, we consider only homogeneous vehicles, with the same vehicle capacity $C$ shared among all vehicles, and the number of vehicles is unlimited. Following the common capacity setup used in previous studies~\cite{kool2018attention,NEURIPS2020_f231f210}, for $N=50$ and $N=100$, the vehicle capacity $C$ is set to 40 and 50 respectively.

\paragraph{Node Demands} In our study, there are two types of customers: linehaul customers with demand $\delta_i < 0$ and backhaul customers with $\delta_i > 0$ (when the backhaul constraint is active). We generate node demands as follows: we generate linehaul demands $\delta_i^l$ for all customers $i \in \{1,\dots, N\}$ by uniform sampling from the set of integers $\{1, 2, ..., 9\}$. If the backhaul constraint is inactive, each node's true demands $\delta_i$ are equal to $\delta_i^l$. The demand generation process is now complete. Otherwise, we generate backhaul demands $\delta_i^b$ by sampling uniformly from the same set of integers $\{1, 2, ..., 9\}$. Subsequently, generate a temporary variable $y_i \sim U(0,1)$ for each customer $i$. The demand $\delta_i$ for each customer $i$ is determined by the following rule:
\begin{equation}
\delta_i = 
\begin{cases} 
    \delta_i^l & \text{if }  y_{i} \ge 0.2, \\
    \delta_i^b & \text{otherwise} .
\end{cases}
\end{equation}
For each node, there is a 20\% probability that it represents backhaul customers in an instance.

Furthermore, before passing the demands $\delta_i$ to the policy, for training stability, we normalize the demand $\delta_i$ to the range $[0, 1]$ by $\delta_i' = \frac{\delta_i}{C}$. We set the normalized capacity to 1 to ensure that at each step of the decoding process, the remaining capacity $c_t$ also falls within the range $[0,1]$.

\paragraph{Time Windows}
For problems with time window constraints, several related factors must be considered: time windows $[e_i, l_i]$ and service times $s_i$. For the depot, $e_0 = s_0 = 0$, $l_0 = \mathcal{T} = 4.6$, where $\mathcal{T}$ represents the overall time limit for each sub-route. Additionally, the vehicle speed is 1.0.

For customers $i \in \{1, 2, \dots, N\}$, service times $s_i$ are uniformly sampled from $[0.15, 0.18]$. Additionally, time window lengths $\Delta t_i$ are uniformly sampled from $[0.18, 0.2]$. Moreover, each customer's time window must be feasible for the tour $(0, i, 0)$; otherwise, there is no feasible tour to service this customer. Consequently, the upper bounds for the start times of the time windows are calculated as:
\begin{equation}
    e_i^{\text{up}} = \frac{\mathcal{T} - s_i - \Delta t_i}{d_{0i}} - 1.
\end{equation}
Subsequently, the start times of the time windows $e_i$ are determined as follows:
\begin{equation}
    e_i = (1 + (e_i^{\text{up}} - 1) \cdot y_i) \cdot d_{0i},
\end{equation}
where $y_i \sim U(0,1)$. Finally, the end times of the time windows are determined by:
\begin{equation}
l_i = e_i + \Delta t_i.
\end{equation}
% When calculating the action mask, we have the constraint that the expected arrival time should be earlier than the end time of nodes; if the problem is a close problem, we should also consider the time back to the depot,  i.e., \(\max(t_{\mathrm{curr}} + d_{ij}, e_j) + s_j + d_{j0} < l_0\). We note that for simplicity, we set the vehicle speed to $1.0$ in equations and normalize time windows accordingly so that travel time from two nodes is the same numerically as the distance between them. This can be easily modified in the code.

\paragraph{Distance Limit}
For problems with the Distance Limit constraint, each sub-tour must be completed within a limit $\rho$. To ensure each instance has a feasible solution, i.e., the length of the tour $(0, i, 0)$ should remain within this limit, $\rho$ is sampled from $U(2 \cdot \max(d_{0*}), \rho_{\text{max}})$, where $\rho_{\text{max}} = 3.0$ is a predefined upper bound.

\begin{table}[t]
\centering
\caption{Experiment hyperparameters.}
\label{tab:hyperparameters}
\renewcommand\arraystretch{1.2}  % 0.97
\begin{tabularx}{\columnwidth}{XX}
\toprule
Hyperparameter& Value \\
\midrule
\multicolumn{2}{l}{\textbf{Model}} \\
Embedding dimension $d_h$ & 128 \\
Number of attention heads $M_h$ & 8 \\
Number of encoder layers $L$ & 6 \\
Top-$k$ & $\frac{(N)}{2} $ \\
% Number of prompts & 5 \\
Feedforward hidden dimension $d_a$ & 512 \\
Tanh clipping $\xi$ & 10.0 \\
\midrule
\multicolumn{2}{l}{\textbf{Training}} \\
% Train decode type & multistart sampling \\
% Test decode type & multistart greedy \\
% Augmentation function & dihedral \\
Batch size & 256 \\
Train data per epoch & 100,000 \\
% \midrule
% \multicolumn{2}{l}{\textbf{Optimization}} \\
Optimizer & AdamW~\cite{loshchilov2017decoupled} \\
Learning rate (LR) & $3\text{e}^{-4}$ \\
Weight decay & $1\text{e}^{-6}$ \\
LR scheduler & MultiStepLR \\
LR milestones & [270, 295] \\
LR gamma & 0.1 \\
Gradient clip value & 1.0 \\
Training epochs & 300 \\
Number of tasks used for training & 16 \\
\bottomrule
\end{tabularx}
\end{table}

\begin{figure}[t]
    \centering
    \subfigure[Average gap of different neural solvers on VRPs with 50 nodes.]{\includegraphics[width = 0.24\textwidth]{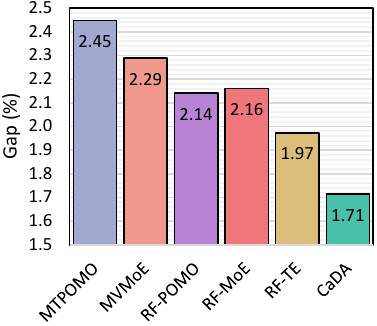}}
    \subfigure[Average gap of different neural solvers on VRPs with 100 nodes.]{\includegraphics[width = 0.24\textwidth]{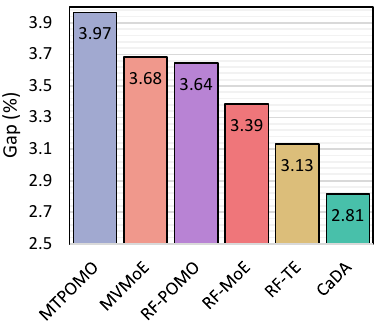}}
    \caption{Comparison results of CaDA with state-of-the-art cross-problem neural solvers, showing the average gap on 16 VRPs.}
    \label{fig:avg16main}
\end{figure}

\begin{table*}[!t]
    \caption{Performance on 1000 test instances of 16 VRPs. The best learning-based results for each dataset are highlighted with a gray background.}
    \label{tab:main-results}
    \begin{center}
    \renewcommand\arraystretch{1.05}
    \resizebox{0.98\textwidth}{!}{ 
    \begin{tabular}{ll|cccccc|ll|cccccc}
      \toprule
      \multicolumn{2}{c|}{\multirow{2}{*}{Solver}} & \multicolumn{3}{c}{\textbf{$n=50$}} & \multicolumn{3}{c|}{$n=100$} & \multicolumn{2}{c|}{\multirow{2}{*}{Solver}} &
      \multicolumn{3}{c}{\textbf{$n=50$}} & \multicolumn{3}{c}{$n=100$} \\
      \cmidrule(lr){3-5} \cmidrule(lr){6-8} \cmidrule(lr){11-13} \cmidrule(lr){14-16}
  
       & & Obj. & Gap & Time & Obj. & Gap & Time & & & Obj. & Gap & Time & Obj. & Gap & Time \\
      \midrule
\multirow{7}*{\rotatebox{90}{CVRP}} 
& HGS-PyVRP & 10.372 & * & 10.4m & 15.628 & * & 20.8m  & \multirow{7}*{\rotatebox{90}{VRPTW}} & HGS-PyVRP & 16.031 & * & 10.4m & 25.423 & * & 20.8m  \\
& OR-Tools & 10.572 & 1.907\% & 10.4m & 16.280 & 4.178\% & 20.8m      & & OR-Tools & 16.089 & 0.347\% & 10.4m & 25.814 & 1.506\% & 20.8m  \\
 & MTPOMO  & 10.520                         & 1.423\%                         & 2s   & 15.941                         & 2.030\%                         & 8s   &  & MTPOMO  & 16.419                         & 2.423\%                         & 2s   & 26.433                         & 3.962\%                         & 9s   \\
 & MVMoE   & 10.499                         & 1.229\%                         & 3s   & 15.888                         & 1.693\%                         & 11s  &  & MVMoE   & 16.400                         & 2.298\%                         & 3s   & 26.390                         & 3.789\%                         & 11s  \\
 & RF-POMO & 10.506                         & 1.300\%                         & 2s   & 15.908                         & 1.833\%                         & 8s   &  & RF-POMO & 16.363                         & 2.066\%                         & 2s   & 26.361                         & 3.675\%                         & 9s   \\
 & RF-MoE  & 10.499                         & 1.225\%                         & 3s   & 15.877                         & 1.625\%                         & 11s  &  & RF-MoE  & 16.389                         & 2.232\%                         & 3s   & 26.321                         & 3.516\%                         & 11s  \\
 & RF-TE   & 10.502                         & 1.257\%                         & 2s   & \cellcolor[HTML]{D9D9D9}15.860 & \cellcolor[HTML]{D9D9D9}1.524\% & 8s   &  & RF-TE   & 16.341                         & 1.933\%                         & 2s   & 26.228                         & 3.154\%                         & 8s   \\
 & CaDA    & \cellcolor[HTML]{D9D9D9}10.494 & \cellcolor[HTML]{D9D9D9}1.182\% & 2s   & 15.870                         & 1.578\%                         & 8s   &  & CaDA    & \cellcolor[HTML]{D9D9D9}16.278 & \cellcolor[HTML]{D9D9D9}1.536\% & 2s   & \cellcolor[HTML]{D9D9D9}26.070 & \cellcolor[HTML]{D9D9D9}2.530\% & 8s   \\
      \midrule
\multirow{7}*{\rotatebox{90}{OVRP}} 
& HGS-PyVRP & 6.507 & * & 10.4m & 9.725 & * & 20.8m  & \multirow{7}*{\rotatebox{90}{VRPL}} & HGS-PyVRP & 10.587 & * & 10.4m & 15.766 & * & 20.8m \\
& OR-Tools & 6.553 & 0.686\% & 10.4m & 9.995 & 2.732\% & 20.8m  & & OR-Tools       & 10.570 & 2.343\% & 10.4m     & 16.466 & 5.302\% & 20.8m     \\
 & MTPOMO  & 6.717                          & 3.194\%                         & 2s   & 10.216                         & 5.028\%                         & 8s   &  & MTPOMO  & 10.775                         & 1.733\%                         & 2s   & 16.157                         & 2.483\%                         & 8s   \\
 & MVMoE   & 6.705                          & 3.003\%                         & 3s   & 10.177                         & 4.617\%                         & 11s  &  & MVMoE   & 10.753                         & 1.525\%                         & 3s   & 16.099                         & 2.113\%                         & 11s  \\
 & RF-POMO & 6.699                          & 2.926\%                         & 2s   & 10.190                         & 4.761\%                         & 8s   &  & RF-POMO & 10.748                         & 1.498\%                         & 2s   & 16.117                         & 2.241\%                         & 8s   \\
 & RF-MoE  & 6.697                          & 2.880\%                         & 3s   & 10.139                         & 4.234\%                         & 11s  &  & RF-MoE  & 10.737                         & 1.390\%                         & 3s   & 16.070                         & 1.937\%                         & 11s  \\
 & RF-TE   & 6.682                          & 2.658\%                         & 2s   & \cellcolor[HTML]{D9D9D9}10.115 & \cellcolor[HTML]{D9D9D9}3.996\% & 8s   &  & RF-TE   & 10.747                         & 1.485\%                         & 2s   & 16.057                         & 1.858\%                         & 8s   \\
 & CaDA    & \cellcolor[HTML]{D9D9D9}6.670  & \cellcolor[HTML]{D9D9D9}2.468\% & 2s   & 10.121                         & 4.045\%                         & 8s   &  & CaDA    & \cellcolor[HTML]{D9D9D9}10.731 & \cellcolor[HTML]{D9D9D9}1.333\% & 2s   & \cellcolor[HTML]{D9D9D9}16.057 & \cellcolor[HTML]{D9D9D9}1.847\% & 8s   \\
      \midrule
\multirow{7}*{\rotatebox{90}{VRPB}} 
& HGS-PyVRP & 9.687 & * & 10.4m & 14.377 & * & 20.8m   & \multirow{7}*{\rotatebox{90}{OVRPTW}} & HGS-PyVRP & 10.510 & * & 10.4m & 16.926 & * & 20.8m   \\
& OR-Tools & 9.802 & 1.159\% & 10.4m & 14.933 & 3.853\% & 20.8m 
& & OR-Tools & 10.519 & 0.078\% & 10.4m & 17.027 & 0.583\% & 20.8m  \\
 & MTPOMO  & 10.036                         & 3.596\%                         & 2s   & 15.102                         & 5.052\%                         & 8s   &  & MTPOMO  & 10.676                         & 1.558\%                         & 2s   & 17.442                         & 3.022\%                         & 9s   \\
 & MVMoE   & 10.007                         & 3.292\%                         & 3s   & 15.023                         & 4.505\%                         & 10s  &  & MVMoE   & 10.674                         & 1.541\%                         & 3s   & 17.416                         & 2.870\%                         & 12s  \\
 & RF-POMO & 9.992                          & 3.135\%                         & 2s   & 15.025                         & 4.534\%                         & 8s   &  & RF-POMO & 10.656                         & 1.361\%                         & 2s   & 17.405                         & 2.809\%                         & 9s   \\
 & RF-MoE  & 9.980                          & 3.017\%                         & 3s   & 14.973                         & 4.168\%                         & 10s  &  & RF-MoE  & 10.674                         & 1.540\%                         & 3s   & 17.388                         & 2.704\%                         & 12s  \\
 & RF-TE   & 9.979                          & 3.000\%                         & 2s   & \cellcolor[HTML]{D9D9D9}14.935 & \cellcolor[HTML]{D9D9D9}3.906\% & 8s   &  & RF-TE   & 10.645                         & 1.264\%                         & 2s   & 17.328                         & 2.352\%                         & 9s   \\
 & CaDA    & \cellcolor[HTML]{D9D9D9}9.960  & \cellcolor[HTML]{D9D9D9}2.800\% & 2s   & 14.960                         & 4.038\%                         & 8s   &  & CaDA    & \cellcolor[HTML]{D9D9D9}10.613 & \cellcolor[HTML]{D9D9D9}0.957\% & 2s   & \cellcolor[HTML]{D9D9D9}17.226 & \cellcolor[HTML]{D9D9D9}1.751\% & 9s   \\

      \midrule
\multirow{7}*{\rotatebox{90}{VRPBL}} 
& HGS-PyVRP & 10.186 & * & 10.4m & 14.779 & * & 20.8m 
& \multirow{7}*{\rotatebox{90}{VRPBLTW}} 
                                                                             &   HGS-PyVRP      & 18.361 & *       & 10.4m     & 29.026 & *       & 20.8m     \\
& OR-Tools & 10.331 & 1.390\% & 10.4m & 15.426 & 4.338\% & 20.8m 
& & OR-Tools       & 18.422 & 0.332\% & 10.4m     & 29.830 & 2.770\% & 20.8m     \\
 & MTPOMO  & 10.679                         & 4.760\%                         & 2s   & 15.718                         & 6.294\%                         & 8s   &  & MTPOMO  & 19.001                         & 2.199\%                         & 3s   & 30.948                         & 3.794\%                         & 9s   \\
 & MVMoE   & 10.639                         & 4.384\%                         & 3s   & 15.642                         & 5.771\%                         & 11s  &  & MVMoE   & 18.983                         & 2.097\%                         & 3s   & 30.892                         & 3.609\%                         & 12s  \\
 & RF-POMO & 10.590                         & 3.926\%                         & 2s   & 15.632                         & 5.725\%                         & 8s   &  & RF-POMO & 18.938                         & 1.863\%                         & 2s   & 30.847                         & 3.452\%                         & 9s   \\
 & RF-MoE  & 10.575                         & 3.765\%                         & 3s   & 15.542                         & 5.125\%                         & 10s  &  & RF-MoE  & 18.957                         & 1.960\%                         & 3s   & 30.809                         & 3.325\%                         & 12s  \\
 & RF-TE   & 10.569                         & 3.713\%                         & 2s   & 15.523                         & 5.008\%                         & 8s   &  & RF-TE   & 18.910                         & 1.713\%                         & 2s   & 30.705                         & 2.978\%                         & 9s   \\
 & CaDA    & \cellcolor[HTML]{D9D9D9}10.543 & \cellcolor[HTML]{D9D9D9}3.461\% & 2s   & \cellcolor[HTML]{D9D9D9}15.525 & \cellcolor[HTML]{D9D9D9}5.001\% & 8s   &  & CaDA    & \cellcolor[HTML]{D9D9D9}18.848 & \cellcolor[HTML]{D9D9D9}1.376\% & 2s   & \cellcolor[HTML]{D9D9D9}30.520 & \cellcolor[HTML]{D9D9D9}2.359\% & 9s   \\
      \midrule
\multirow{7}*{\rotatebox{90}{VRPBTW}} 
& HGS-PyVRP & 18.292 & * & 10.4m & 29.467 & * & 20.8m & \multirow{7}*{\rotatebox{90}{VRPLTW}} & HGS-PyVRP & 16.356 & * & 10.4m & 25.757 & * & 20.8m \\
& OR-Tools & 18.366 & 0.383\% & 10.4m & 29.945 & 1.597\% & 20.8m & & OR-Tools & 16.441 & 0.499\% & 10.4m & 26.259 & 1.899\% & 20.8m    \\
 & MTPOMO  & 18.649                         & 1.938\%                         & 2s   & 30.478                         & 3.426\%                         & 9s   &  & MTPOMO  & 16.832                         & 2.877\%                         & 2s   & 26.913                         & 4.455\%                         & 9s   \\
 & MVMoE   & 18.632                         & 1.841\%                         & 3s   & 30.437                         & 3.284\%                         & 12s  &  & MVMoE   & 16.817                         & 2.783\%                         & 3s   & 26.866                         & 4.272\%                         & 12s  \\
 & RF-POMO & 18.603                         & 1.684\%                         & 2s   & 30.384                         & 3.102\%                         & 9s   &  & RF-POMO & 16.756                         & 2.419\%                         & 2s   & 26.818                         & 4.084\%                         & 9s   \\
 & RF-MoE  & 18.616                         & 1.757\%                         & 3s   & 30.340                         & 2.951\%                         & 12s  &  & RF-MoE  & 16.777                         & 2.548\%                         & 3s   & 26.773                         & 3.910\%                         & 12s  \\
 & RF-TE   & 18.573                         & 1.517\%                         & 2s   & 30.249                         & 2.641\%                         & 9s   &  & RF-TE   & 16.728                         & 2.248\%                         & 2s   & 26.706                         & 3.645\%                         & 9s   \\
 & CaDA    & \cellcolor[HTML]{D9D9D9}18.500 & \cellcolor[HTML]{D9D9D9}1.117\% & 2s   & \cellcolor[HTML]{D9D9D9}30.059 & \cellcolor[HTML]{D9D9D9}1.999\% & 9s   &  & CaDA    & \cellcolor[HTML]{D9D9D9}16.669 & \cellcolor[HTML]{D9D9D9}1.879\% & 2s   & \cellcolor[HTML]{D9D9D9}26.540 & \cellcolor[HTML]{D9D9D9}2.995\% & 9s   \\
      \midrule
\multirow{7}*{\rotatebox{90}{OVRPB}} 
& HGS-PyVRP & 6.898 & * & 10.4m & 10.335 & * & 20.8m  & \multirow{7}*{\rotatebox{90}{OVRPBL}} & HGS-PyVRP & 6.899 & * & 10.4m & 10.335 & * & 20.8m \\
& OR-Tools & 6.928 & 0.412\% & 10.4m & 10.577 & 2.315\% & 20.8m 
& & OR-Tools & 6.927 & 0.386\% & 10.4m & 10.582 & 2.363\% & 20.8m  \\
 & MTPOMO  & 7.105                          & 2.973\%                         & 2s   & 10.882                         & 5.264\%                         & 8s   &  & MTPOMO  & 7.112                          & 3.053\%                         & 2s   & 10.888                         & 5.318\%                         & 8s   \\
 & MVMoE   & 7.089                          & 2.744\%                         & 3s   & 10.841                         & 4.869\%                         & 11s  &  & MVMoE   & 7.094                          & 2.799\%                         & 3s   & 10.847                         & 4.929\%                         & 11s  \\
 & RF-POMO & 7.085                          & 2.686\%                         & 2s   & 10.839                         & 4.857\%                         & 8s   &  & RF-POMO & 7.088                          & 2.703\%                         & 2s   & 10.842                         & 4.883\%                         & 8s   \\
 & RF-MoE  & 7.081                          & 2.617\%                         & 3s   & 10.806                         & 4.528\%                         & 11s  &  & RF-MoE  & 7.082                          & 2.630\%                         & 3s   & 10.807                         & 4.537\%                         & 11s  \\
 & RF-TE   & 7.065                          & 2.385\%                         & 2s   & 10.774                         & 4.233\%                         & 8s   &  & RF-TE   & 7.068                          & 2.417\%                         & 2s   & 10.778                         & 4.266\%                         & 8s   \\
 & CaDA    & \cellcolor[HTML]{D9D9D9}7.049  & \cellcolor[HTML]{D9D9D9}2.159\% & 2s   & \cellcolor[HTML]{D9D9D9}10.762 & \cellcolor[HTML]{D9D9D9}4.099\% & 8s   &  & CaDA    & \cellcolor[HTML]{D9D9D9}7.051  & \cellcolor[HTML]{D9D9D9}2.166\% & 2s   & \cellcolor[HTML]{D9D9D9}10.762 & \cellcolor[HTML]{D9D9D9}4.102\% & 8s   \\
      \midrule
\multirow{7}*{\rotatebox{90}{OVRPBLTW}} & HGS-PyVRP & 11.668 & * & 10.4m & 19.156 & * & 20.8m 
& \multirow{7}*{\rotatebox{90}{OVRPBTW}} & HGS-PyVRP & 11.669 & * & 10.4m & 19.156 & * & 20.8m  \\
& OR-Tools & 11.681 & 0.106\% & 10.4m & 19.305 & 0.767\% & 20.8m & & OR-Tools & 11.682 & 0.109\% & 10.4m & 19.303 & 0.757\% & 20.8m  \\
 & MTPOMO  & 11.823                         & 1.315\%                         & 3s   & 19.658                         & 2.602\%                         & 9s   &  & MTPOMO  & 11.823                         & 1.307\%                         & 3s   & 19.656                         & 2.592\%                         & 9s   \\
 & MVMoE   & 11.816                         & 1.249\%                         & 4s   & 19.640                         & 2.514\%                         & 12s  &  & MVMoE   & 11.816                         & 1.245\%                         & 4s   & 19.637                         & 2.499\%                         & 13s  \\
 & RF-POMO & 11.810                         & 1.192\%                         & 3s   & 19.618                         & 2.393\%                         & 10s  &  & RF-POMO & 11.809                         & 1.182\%                         & 3s   & 19.620                         & 2.403\%                         & 10s  \\
 & RF-MoE  & 11.824                         & 1.309\%                         & 4s   & 19.607                         & 2.334\%                         & 12s  &  & RF-MoE  & 11.823                         & 1.303\%                         & 4s   & 19.605                         & 2.324\%                         & 12s  \\
 & RF-TE   & 11.789                         & 1.017\%                         & 2s   & 19.554                         & 2.061\%                         & 9s   &  & RF-TE   & 11.790                         & 1.027\%                         & 2s   & 19.555                         & 2.062\%                         & 9s   \\
 & CaDA    & \cellcolor[HTML]{D9D9D9}11.760 & \cellcolor[HTML]{D9D9D9}0.771\% & 2s   & \cellcolor[HTML]{D9D9D9}19.435 & \cellcolor[HTML]{D9D9D9}1.439\% & 9s   &  & CaDA    & \cellcolor[HTML]{D9D9D9}11.761 & \cellcolor[HTML]{D9D9D9}0.779\% & 2s   & \cellcolor[HTML]{D9D9D9}19.436 & \cellcolor[HTML]{D9D9D9}1.441\% & 9s   \\

\midrule
\multirow{7}*{\rotatebox{90}{OVRPL}} 
& HGS-PyVRP & 6.507 & * & 10.4m & 9.724 & * & 20.8m 
 & \multirow{7}*{\rotatebox{90}{OVRPLTW}} & HGS-PyVRP & 10.510 & * & 10.4m & 16.926 & * & 20.8m \\
& OR-Tools & 6.552 & 0.668\% & 10.4m & 10.001 & 2.791\% & 20.8m  & & OR-Tools       & 10.497 & 0.114\% & 10.4m     & 17.023 & 0.728\% & 20.8m     \\
 & MTPOMO  & 6.720                          & 3.248\%                         & 2s   & 10.224                         & 5.112\%                         & 8s   &  & MTPOMO  & 10.677                         & 1.572\%                         & 2s   & 17.442                         & 3.020\%                         & 9s   \\
 & MVMoE   & 6.706                          & 3.028\%                         & 3s   & 10.184                         & 4.693\%                         & 11s  &  & MVMoE   & 10.677                         & 1.564\%                         & 3s   & 17.418                         & 2.880\%                         & 12s  \\
 & RF-POMO & 6.701                          & 2.944\%                         & 2s   & 10.190                         & 4.762\%                         & 8s   &  & RF-POMO & 10.656                         & 1.362\%                         & 3s   & 17.404                         & 2.802\%                         & 9s   \\
 & RF-MoE  & 6.695                          & 2.859\%                         & 3s   & 10.140                         & 4.252\%                         & 11s  &  & RF-MoE  & 10.673                         & 1.531\%                         & 3s   & 17.386                         & 2.696\%                         & 12s  \\
 & RF-TE   & 6.683                          & 2.680\%                         & 2s   & 10.121                         & 4.054\%                         & 8s   &  & RF-TE   & 10.646                         & 1.267\%                         & 2s   & 17.328                         & 2.352\%                         & 9s   \\
 & CaDA    & \cellcolor[HTML]{D9D9D9}6.671  & \cellcolor[HTML]{D9D9D9}2.475\% & 2s   & \cellcolor[HTML]{D9D9D9}10.122 & \cellcolor[HTML]{D9D9D9}4.052\% & 8s   &  & CaDA    & \cellcolor[HTML]{D9D9D9}10.613 & \cellcolor[HTML]{D9D9D9}0.961\% & 2s   & \cellcolor[HTML]{D9D9D9}17.226 & \cellcolor[HTML]{D9D9D9}1.752\% & 9s   \\
      \bottomrule
    \end{tabular}}
    \end{center}
  \end{table*}

\subsection{Baselines} We utilize state-of-the-art traditional and neural solvers as baselines. 
For the traditional approaches, we employ two open-source solvers capable of addressing all 16 VRPs in this study: PyVRP~\cite{wouda2024pyvrp}, which extends the state-of-the-art heuristic algorithm HGS-CVRP~\cite{vidal2022hybrid}, and Google's OR-Tools. Both baseline methods solve each instance using a single CPU core, with a time limit of 10 seconds for VRP50 and 20 seconds for VRP100, respectively.
For the neural solvers, we compare our method against representative multi-task learning models: MTPOMO~\cite{2024arXiv240216891L}, MVMoE~\cite{zhou2024mvmoe}, and RouteFinder~\cite{berto2024routefinder}, including RF-POMO, RF-MoE, and RF-TE. We utilize the open-source code published by RouteFinder~\cite{berto2024routefinder}. For each method, we separately train two models from scratch on VRP50 and VRP100 using the same hyperparameter settings in RouteFinder~\cite{berto2024routefinder}. However, for RF-MoE and MVMoE, due to their higher memory demands, we utilize the pre-trained parameters provided by RouteFinder~\cite{berto2024routefinder} and test them under the same hardware settings as ours. For our method CaDA, we also train two models on VRP50 and VRP100, with the hyperparameters summarized in Table~\ref{tab:hyperparameters}. All neural methods utilize the same training data budget.

\subsection{Testing and Hardware} We utilize the test dataset published by Routefinder~\cite{berto2024routefinder}, which includes 1K randomly generated instances for each VRP variant, at scales of 50 and 100. All neural solvers are tested on the same scale as their training scale. For all neural solvers, we employ a greedy rollout strategy with $\times 8$aug~\cite{kool2018attention}. This approach conducts equivalent transformations to augment the original instance and reports the best results among the eight augmented instances.

All computational experiments are conducted on a platform equipped with NVIDIA GeForce RTX 3090 GPUs and Intel(R) Xeon(R) Gold 6348 CPUs at 2.60 GHz. Under these conditions, training our model from scratch for VRP50 takes approximately 17 hours, and about 25 hours for VRP100.

\begin{table}[t]
\caption{Average objective function value and gap across 16 VRPs for ablation models and CaDA.}
\label{tab:ablation}
\centering
\renewcommand\arraystretch{1.2}  % 0.97
\begin{tabularx}{\columnwidth}{l|XX}
\toprule
& Obj. & Gap \\
\midrule
CaDA w/o Prompt         & 11.534          & 1.926\%          \\
CaDA w/o Sparse & 11.521          & 1.795\%          \\
CaDA               & \textbf{11.513} & \textbf{1.714\%} \\
\bottomrule
\end{tabularx}
\end{table}

\begin{figure}[t]
\centering
\includegraphics[width=0.99\linewidth]{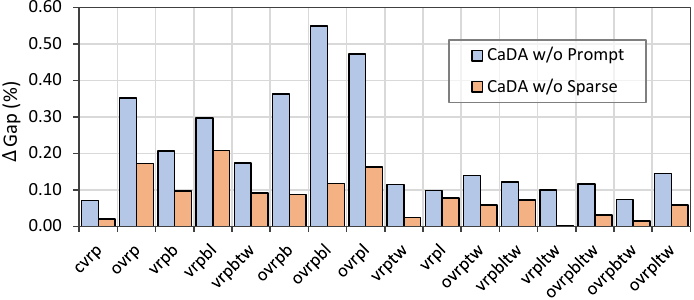}
\caption{Ablation study on the proposed components of CaDA across 16 VRPs. The height of the bars represents the increased gap in performance between the model with specific components ablated and the baseline CaDA.}
\label{fig:ablationp}
\end{figure} 

\subsection{Main Results}
In Table~\ref{tab:main-results}, we report the average performances for each dataset and the gaps compared to the best-performing traditional VRP solvers, as indicated by asterisks (*). Additionally, we present the total time required to solve the dataset. The best learning-based results for each dataset are highlighted with a gray background. Furthermore, we present the comparison results with the average gap on 16 VRPs for both 50-node and 100-node instances in Figure \ref{fig:avg16main}.

% Results demonstrate that the proposed CaDA can efficiently handle multiple VRP variants jointly in both scales. For VRP50 and VRP100, CaDA outperforms the second-best method by 0.26\% and 0.32\%, respectively. It ranks first among all the neural solvers on all VRP50 variants and 13 out of 16 on VRP100 variants. Similar to other neural solvers, the running time is significantly reduced when compared to state-of-the-art heuristic solvers.

Results illustrate that the proposed CaDA method can effectively manage various VRP variants at different scales. Specifically, for VRP50 and VRP100, CaDA surpasses the second-best method by 0.26\% and 0.32\%, respectively. It ranks first among all neural network-based solvers for all VRP50 variants and for 13 out of 16 VRP100 variants. Similar to other neural solvers, CaDA significantly reduces the running time compared to state-of-the-art heuristic solvers.

\subsection{Ablation Study}

In this section, we conduct ablation studies to validate the efficacy of the proposed components in CaDA. Specifically, we separately remove the constraint prompt and the Top-$k$ operation, resulting in two CaDA variants: CaDA w/o prompt and CaDA w/o Sparse, where 'w/o' stands for 'without.'  In CaDA w/o Sparse, both branches use the standard Softmax with global connectivity. CaDA and its variants are trained and evaluated on VRP50. During testing, $\times 8$aug~\cite{kool2018attention} is employed.

The results in Table~\ref{tab:ablation} show the average gap on 16 VRPs and demonstrate that all components of CaDA make substantial contributions, with the prompt playing a particularly important role. Additionally, to study the effect of different components on various problems, we demonstrate the increased gap of these two ablation models compared with CaDA on different VRP datasets in Figure~\ref{fig:ablationp}. These results illustrate that both the prompt and the sparse operation improve performance on all VRP variants.
% These results illustrate that the prompt significantly improves performance on problems with the Open Route constraint, such as OVRP, OVRPB, OVRPL, OVRPBL, while the sparse operation brings substantial enhancements in VRPBL compared to other problems.

% Table~\ref{tab:ablation} shows the average gap on 16 VRPs, and Figure ~\ref{fig:ablation} illustrates the convergence process of the different models. The results indicate that all components of CaDA make substantial contributions, with the prompt playing a particularly important role. Additionally, CaDA demonstrates significant advantages in convergence speed and stability.

% \begin{figure}[t]
% \centering
% \includegraphics[width=0.99\linewidth]{figures/ablationfurther.pdf}
% \caption{.
% }
% \label{fig:ablation}
% \end{figure} 

\begin{figure}[t]
    \centering
    \subfigure[CaDA with different prompt positions: concatenated to the input of the sparse branch, and concatenated to the input of the global branch, where the latter is the standard setting.]{\includegraphics[width = 0.24\textwidth]{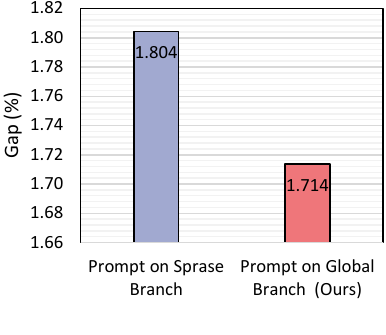}}
    \subfigure[CaDA with different sparse operations in the sparse branch, and without sparse operations (i.e., using only standard softmax in both branches), where Top-$k$ is the standard setting.]{\includegraphics[width = 0.24\textwidth]{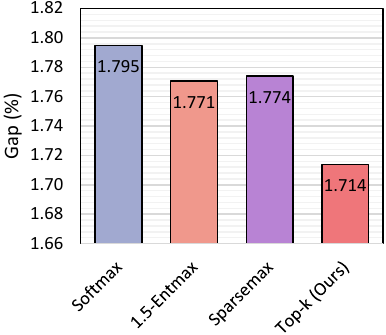}}
    \caption{The average gap across 16 VRPs for CaDA variants under different model settings.}
    \label{fig:curve}
\end{figure}

\begin{figure}[t]
\centering
\includegraphics[width=0.9\linewidth]{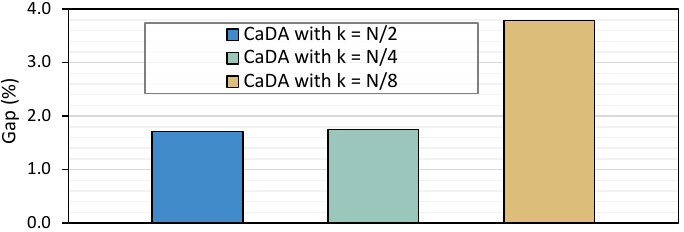}
\caption{Average gap on 16 VRPs for CaDA variants with different $k$ values in the Top-$k$ selection operation, where $k = \frac{N}{2}$ is the standard setting.}
\label{fig:ablationk}
\end{figure} 

\begin{figure*}[t]
\centering
\includegraphics[width=1.0\linewidth]{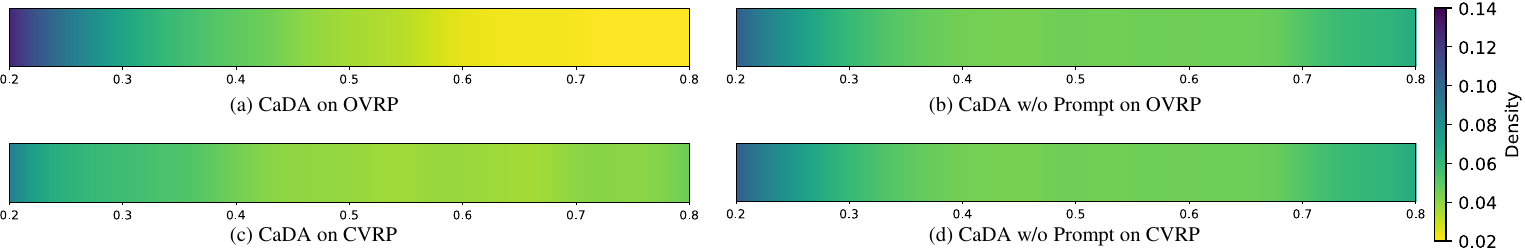}
\caption{The distribution of attention weights between customers and the depot $\mathbf{A}_{i0}, i \in \{ 1,2, \dots, N\}$ for CaDA and CaDA w/o Prompt on CVRP and OVRP. Kernel Density Estimation (KDE) with Gaussian kernels is applied to estimate the attention weight distribution, which is visualized using a heatmap. CaDA w/o Prompt exhibits a similar attention distribution across both problems, leading to increased interference between tasks. In contrast, CaDA shows a significantly lower density of high attention values for OVRP, indicating that the proposed prompt effectively provides constraint information as depot will never be the next action for any customer in OVRP.}
\label{fig:attnbydiso}
\end{figure*}

\begin{figure}[t]
    \centering
    \subfigure[CaDA]{\includegraphics[width = 0.24\textwidth]{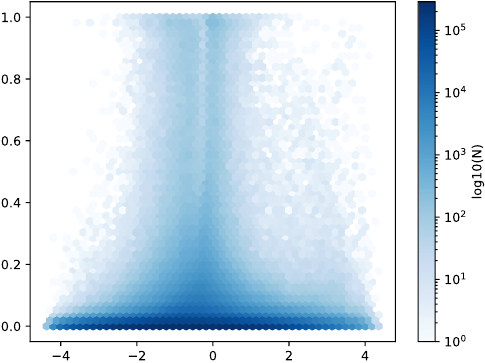}}    
    \subfigure[CaDA w/o Prompt]{\includegraphics[width = 0.24\textwidth]{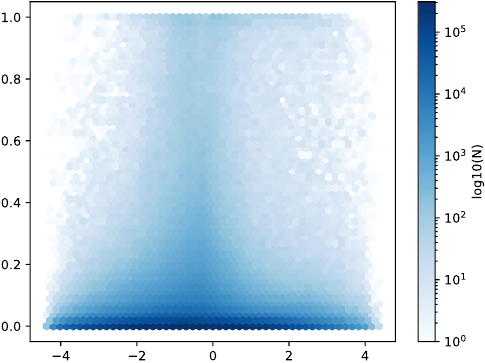}}    
    % \subfigure[CaDA w/o Sparse]{\includegraphics[width = 0.32\textwidth]{figures/AttnbyDisBin_6.pdf}}
    \caption{
    For 100 VRPTW instances, the distribution of attention scores $\mathbf{A}_{ij}$ across varying $\mathbf{P}_{i,j}$, where $\mathbf{A}_{ij}$ is the attention score from node $v_i$ to node $v_j$, and $\mathbf{P}_{i,j} = (l_j - e_i) - d_{ij} - (s_i + s_j)$ represents the surplus time when the vehicle starts from $v_i$ at time $e_i$ and travels to $v_j$. If $\mathbf{P}_{i,j} < 0$, the edge $(i,j)$ is illegal; if $\mathbf{P}_{i,j}$ becomes too large, including $(i,j)$ in the solution may result in the vehicle having to wait a long time for $v_j$ to open. The shade represents the density in that region. For CaDA, the attention scores at extremely high and low $\mathbf{P}_{i,j}$ values are diminished, indicating that the model successfully comprehends the time window constraint.
    }
    \label{fig:attnbydistw}
\end{figure}

Then, we further explore the influence of different CaDA settings. Specifically, using CaDA as the baseline, we conduct experiments on VRP50 focusing on the following aspects:

\paragraph{Position of Prompt} We consider two positions to introduce prompts: the input for the global branch (i.e., CaDA), and the input for the sparse branch. Figure~\ref{fig:curve}(a) illustrates the average gap across 16 VRPs of the different models. The results indicate that integrating both sparse and prompt mechanisms within the same branch yields inferior performance compared to placing the prompt in the global branch. This is because the prompt in the sparse branch may be masked by the sparse function, leading to insufficient constraint information, which diminishes model performance.

\paragraph{Different Sparse Function} To validate the efficacy of our Top-$k$ sparse operation, we replace the Softmax+Top-$k$ in our CaDA with a standard Softmax and a representative sparse function $\alpha$-entmax with two configurations: $\alpha = 1.5$, referred to as 1.5-entmax~\cite{peters2019entmax}, and $\alpha = 2$, which corresponds to sparsemax~\cite{martins2016sparsemax}. 

Figure ~\ref{fig:curve}(b) shows the average gap across 16 VRPs. Firstly, three versions of CaDA that incorporate different sparse functions consistently outperform the version of CaDA with standard Softmax, which only has global connectivity. This indicates that the model benefits from focusing on promising nodes. Furthermore, the Top-$k$ operation outperforms the other two sparse operations, demonstrating its effectiveness for VRPs.

% \begin{figure*}[t]
%     \centering
%     \subfigure[CaDA]{\includegraphics[width = 0.32\textwidth]{figures/AttnbyDisBin_1.pdf}}    
%     \subfigure[CaDA w/o Prompt]{\includegraphics[width = 0.32\textwidth]{figures/AttnbyDisBin_2.pdf}}    
%     \subfigure[CaDA w/o Sparse]{\includegraphics[width = 0.32\textwidth]{figures/AttnbyDisBin_3.pdf}}    
%     \caption{
%     % Distribution of attention scores across varying distances in OVRP.
%     For 100 OVRP instances, the distribution of attention scores $\mathbf{A}_{i0}$ across varying distances $d_{i0}$, where $\mathbf{A}_{i0}$ is the attention score from customer $v_i$ to the depot $v_0$, and $d_{i0}$ is the corresponding distance between $v_i$ and $v_0$. The shade represents the density in that region.  With the proposed sparse operation and constraint prompt, the attention scores are concentrated in lower value regions, indicating that both help the model understand the open route constraint, where the depot will never be the next node for any customer.
%     }
%     \label{fig:attnbydiso}
% \end{figure*}

\paragraph{$k$ for Top-$k$} To explore the different $k$ values' effects on CaDA, we conducted experiments with $k \in \left\{\frac{N}{2}, \frac{N}{4}, \frac{N}{8}\right\}$ and trained these models from scratch respectively. Figure~\ref{fig:ablationk} shows the average gap on 16 VRP50 datasets. While $k = \frac{N}{2}$ demonstrates the best performance, $k = \frac{N}{4}$ performs slightly worse, and the performance of $k = \frac{N}{8}$ drops dramatically. It indicates that when $k = \frac{N}{2}$, the sparse branch can better cooperate with the global branch. This is the standard setting for CaDA.

\subsection{Visualization of Constraint Awareness}
% Given the encoding-decoding framework of the NCO model, the encoder outputs node embeddings which the decoder then uses to build the solution one by one. During each decoding step, the current node embedding is processed through a simple glimpse function and then directly used to calculate dot product attention similarity with feasible node embeddings. Thus, the degree of similarity between two node embeddings significantly determines whether one will be the next node of another in the decoding process. Consequently, during the encoding process, blocks with attention functions should aim to bring nodes that may be adjacent in feasible and superior solutions closer together, while disregarding other irrelevant nodes, by efficiently allocating attention scores.

% From this perspective, We conducted further statistical experiments to explore the influence of our components on the attention score distribution within the encoder on VRP50. Both experiments mentioned below involve randomly selecting 100 instances from the test dataset, and attention scores are collected from all heads across all global layers. Additionally, we use the hexbin plot for statistical analysis, where the plane is divided into small hexagons. The color of each hexagon represents the count of data points within that bin, with darker colors indicating higher densities and lighter colors indicating lower densities.

To explore the influence of the prompt, we conducted further statistical experiments on the distribution of attention scores within the encoder. All experiments mentioned below involve 100 VRP50 instances randomly selected from the test dataset, and attention scores were collected from all heads across all global layers.

\begin{table}[t]
\caption{Average objective function value and gap across 16 VRPs for ablation models and CaDA.}
\label{tab:libk}
\centering
\renewcommand\arraystretch{1.2}  % 0.97
\begin{tabularx}{\columnwidth}{l|XX}
\toprule
$k$ & Obj.            & Gap             \\
\midrule
10  & 6440.8          & 4.61\%          \\
25  & \textbf{6431.8} & \textbf{4.17\%} \\
50  & 6458.6          & 4.79\%          \\
75  & 6463.7          & 4.88\%          \\
100 & 6475.7          & 5.15\%          \\
\bottomrule
\end{tabularx}
\end{table}

\begin{table*}[t]
\centering
\caption{Results on CVRPLib datasets.}
\renewcommand\arraystretch{1.3}  % 0.97
  \resizebox{1.0\textwidth}{!}{ 
\begin{tabular}{lc|cccccccccccccc}
    \toprule
 &  & \multicolumn{2}{c}{MTPOMO} & \multicolumn{2}{c}{MVMoE} & \multicolumn{2}{c}{RF-POMO} & \multicolumn{2}{c}{RF-MoE} & \multicolumn{2}{c}{RF-TE} & \multicolumn{2}{c}{$\text{CaDA}_{25}$} & \multicolumn{2}{c}{$\text{CaDA}_{25}\times \text{32}$}      \\
      & Opt.    & Obj.    & Gap     & Obj.    & Gap     & Obj.    & Gap     & Obj.    & Gap     & Obj.            & Gap             & Obj.    & Gap    & Obj.             & Gap             \\
    \midrule
Set A & 1041.9  & 1087.9  & 5.07\%  & 1071.3  & 3.07\%  & 1064.1  & 2.11\%  & 1072.2  & 2.83\%  & 1070.3          & 2.86\%          & 1069.9  & 2.76\% & \textbf{1062.9}  & \textbf{2.00\%} \\
Set B & 963.7   & 1006.9  & 4.86\%  & 999.2   & 3.94\%  & 991.6   & 2.89\%  & 994.0   & 3.17\%  & 987.4           & 2.58\%          & 987.5   & 2.58\% & \textbf{982.7}   & \textbf{2.02\%} \\
Set F & 707.7   & 820.0   & 16.23\% & 770.5   & 7.38\%  & 804.0   & 13.93\% & 813.0   & 14.31\% & 794.7           & 12.95\%         & 748.7   & 5.74\% & \textbf{745.3}   & \textbf{4.99\%} \\
Set P & 587.4   & 629.3   & 11.10\% & 1144.0  & 5.31\%  & 606.4   & 4.72\%  & 603.3   & 3.39\%  & 608.7           & 4.59\%          & 607.9   & 4.82\% & \textbf{601.2}   & \textbf{3.16\%} \\
Set X & 27220.1 & 28952.5 & 6.09\%  & 614.0   & 6.76\%  & 28825.4 & 5.54\%  & 29125.2 & 6.37\%  & 28520.3         & 4.46\%          & 28745.0 & 4.97\% & \textbf{28573.9} & \textbf{4.41\%} \\
\midrule
Avg. & 6104.2 & 6499.3 & 8.67\% & 919.8 & 5.29\% & 6458.3 & 5.84\% & 6521.5 & 6.01\% & 6396.3 & 5.49\% & 6431.8 & 4.17\% & \textbf{6393.2} & \textbf{3.32\%} \\
    \bottomrule
\end{tabular}
}
\label{tab:lib}
\end{table*}

\paragraph{Influence on Open Route Constraint}

In the case of the Open Route constraint, where the vehicle does not return to the depot $v_0$, the depot will never be the next node for any customer node $v_i$. Whereas in CVRP, the vehicle must return to the depot, making it a potential next node for any customer. As a result, the model should exhibit different customer-depot attention patterns for CVRP and OVRP. Specifically, in CVRP, the model should exhibit a greater number of high attention scores $\mathbf{A}_{i0}$ compared to OVRP.

Figure~\ref{fig:attnbydiso} shows the distribution of $\mathbf{A}_{i0}$ for CaDA and CaDA w/o Prompt on CVRP and OVRP. We apply Kernel Density Estimation (KDE) with Gaussian kernels to estimate the distribution of attention weights between 0.2 and 0.8. The bandwidth parameter is set to 0.1, and the KDE is visualized using a heatmap. Firstly, CaDA exhibits a significantly different distribution of attention scores between CVRP and OVRP, whereas CaDA w/o Prompt shows a similar attention distribution across both problems, which increases the interference between different tasks. Furthermore, when comparing CaDA on CVRP and CaDA on OVRP, we observe a significantly lower density of high attention values for OVRP, indicating that the proposed prompt effectively provides constraint information and helps the model better understand the problem.

\paragraph{Influence on Time Window Constraint}
For the Time Window constraint, nodes $v_i$ can only be visited within their respective time windows $(e_i, l_i)$, and serving each customer costs $s_i$ time. Thus, the relationship between node pairs $(v_i, v_j)$ is influenced by their time-related factors and positional distances, i.e., $(e_i, l_i, s_i)$, $(e_j, l_j, s_j)$, and $d_{ij}$.

If an edge $(i, j)$ is legal, then it must satisfy $e_i + s_i + d_{ij} \leq l_j - s_j$. This condition means that if a vehicle starts at time $e_i$, spends $s_i$ time servicing $v_i$, and takes $d_{ij}$ time to reach $v_j$, it must arrive by $(l_j - s_j)$ at the latest to successfully service $v_j$ and leave the node within the service window. Accordingly, we derive the following inequality:
\begin{equation}
    \begin{aligned}
    & e_i + s_i + d_{ij} & \leq l_j - s_j \\
     % \Longrightarrow \quad & l_j - s_j - e_i - s_i - d(i, j) & \geq 0 \\
     \Longrightarrow \quad & (l_j - e_i) - d_{ij} - (s_i + s_j) & \geq 0 
    \end{aligned}
\end{equation}

Define $\mathbf{P}_{i, j} = (l_j - e_i) - d_{ij} - (s_i + s_j)$. Figure \ref{fig:attnbydistw} visualizes the distribution of $\mathbf{A}_{ij}$ across varying $\mathbf{P}_{ij}$. 
% Compared to two ablation models, CaDA exhibits fewer high attention values for $\mathbf{P}{i,j} < 0$. Additionally, for $\mathbf{P}_{ij}$ values that are too high, the vehicle may need to wait a long time at $v_j$ to wait until the start of the time window $e_j$. For example, consider $\mathbf{P}_{ij} = 4$ while the overall time limit for the sub-route is $\mathcal{T} = 4.6$. Including the edge $(i,j)$ in the solution may result in significant time wasted waiting, making it difficult for the vehicle to return to $v_0$. Compared to the ablation models, CaDA also exhibits fewer high attention values for $\mathbf{P}_{ij} > 3$, indicating that the proposed components are effective in focusing attention on more promising nodes.
Firstly, CaDA exhibits fewer high attention values for $\mathbf{P}{i,j} < 0$. Additionally, for $\mathbf{P}_{ij}$ values that are too high, the vehicle may need to wait a long time at $v_j$ to wait until the start of the time window $e_j$. For example, consider $\mathbf{P}_{ij} = 4$ while the overall time limit for the sub-route is $\mathcal{T} = 4.6$. Including the edge $(i,j)$ in the solution may result in significant time wasted waiting. Compared to the CaDA w/o Prompt, CaDA also exhibits fewer high attention values for $\mathbf{P}_{ij} > 3$, indicating that CaDA more efficiently understands the problem constraints.

\subsection{Result on Real-World Instances}
To further validate the effectiveness of CaDA in real-world instances, we conducted experiments using sive test suites from CVRPLib\footnote{\url{http://vrp.atd-lab.inf.puc-rio.br/}} benchmark datasets. These datasets comprise a total of 101 instances from Sets A, B, F, P, and X \cite{uchoa2017new}, with graph scales ranging from 16 to 200, various node distributions, and customer demands. CaDA is trained on 16 different VRP types, each with a graph size of 100 nodes.

We initially explore the effect of different $k$ values on the Top-$k$ sparse operation during testing. Table \ref{tab:libk} demonstrates the average gap compared to the best-known results from CVRPLIB across all instances when testing CaDA with different $k \in \{10, 25, 50, 100\}$. 
% For instances where the graph size is less than $k$, we use $k = N$. 
The results show that when $k=25$, CaDA achieves the best performance across different scales. 
% Conversely, a decrease in sparsity (with $k \geq 50$) correlates with a deterioration in performance, indicating that irrelevant items may adversely affect the performance of the model.

Furthermore, we compared CaDA with existing cross-problem neural solvers. We additionally provide a $\times 32$ data augmentation method for CaDA, which reports the best solution among utilizing 32 prompts. Specifically, we generate all possible combinations of 5-dimensional binary vector $V$ (i.e., $2^5$), which is simple to implement and efficient to execute in batches. Table \ref{tab:lib} exhibits the comparison results of five test suites with the average objective function values, the gap with the best-known solution for each dataset. The best results for each dataset are highlighted in bold. $\text{CaDA}_{25}$ represents CaDA tested with $k = 25$. Results demonstrate that the proposed $\text{CaDA}_{25} \times \text{32}$ achieves the best performance among the learning methods. 
% Additionally, when comparing $\text{CaDA}{25}$ with the prompt of Capacity constraint and $\text{CaDA}{25} \times \text{32 pre}$, which is augmented by testing on all possible prompts, the latter achieves significantly better performance. The CVRPLib datasets include CVRP with varying node distributions, and the prompts for CaDA are only trained on a uniform distribution. Thus, the constraints' prompts may not align with other distributions, while $\times \text{32 pre}$ provides a batchable method to search for promising prompts.

\section{Conclusion}
In this paper, we have proposed the Constraint-Aware Dual-Attention Model (CaDA), a novel cross-problem neural solver for VRPs. CaDA incorporates a constraint prompt and a dual-attention mechanism , which consists of a global branch and a sparse branch, to efficiently generate constraint-aware node embeddings. We have thoroughly evaluated CaDA across 16 VRP variants and real-world benchmark instances. CaDA shows superior performance when compared to current leading neural solvers. Additional ablation studies confirm the effectiveness of the proposed constraint prompt and the dual-attention model structure.

% \textbf{Limitation and Future Work.}   Although the sparse attention introduced to CaDA has enhanced its capability to extract representations, the selection of $k$ for the Top-$k$ mechanism introduces a new hyperparameter that needs to be manually adjusted. Considering that the scale of the feasible space may differ between different constraints, future work could explore a learnable $k$ value for different problems. 

% \section*{Acknowledgment}

% The preferred spelling of the word ``acknowledgment'' in America is without 
% an ``e'' after the ``g''. Avoid the stilted expression ``one of us (R. B. 
% G.) thanks $\ldots$''. Instead, try ``R. B. G. thanks$\ldots$''. Put sponsor 
% acknowledgments in the unnumbered footnote on the first page.

% \section*{References}

\bibliographystyle{IEEEtran}
\bibliography{IEEEabrv,ref}

\end{document}